\documentclass{article}

\usepackage{microtype}
\usepackage{graphicx}
\usepackage{subcaption}
\usepackage{booktabs} 
\usepackage{multirow}

\usepackage{enumitem}
\usepackage{microtype}
\usepackage{hyperref}
\usepackage[most]{tcolorbox}
\newtcbox{\numstep}{
  on line,
  colback=blue!30,  
  colframe=blue!80, 
  boxrule=0pt,      
  arc=0pt,          
  boxsep=0pt,       
  left=2pt, right=2pt, top=1pt, bottom=1pt 
}
\tcbuselibrary{raster}

\usepackage[preprint]{icml2026}

\usepackage{amsmath}
\usepackage{amssymb}
\usepackage{mathtools}
\usepackage{amsthm}

\usepackage[capitalize,noabbrev]{cleveref}

\usepackage{amsmath,amsfonts,bm}

\def\eqref#1{equation~\ref{#1}}

\def\1{\bm{1}}

\def\vd{{\bm{d}}}
\def\ve{{\bm{e}}}

\def\vp{{\bm{p}}}

\def\mE{{\bm{E}}}

\def\mM{{\bm{M}}}

\def\mW{{\bm{W}}}

\DeclareMathAlphabet{\mathsfit}{\encodingdefault}{\sfdefault}{m}{sl}
\SetMathAlphabet{\mathsfit}{bold}{\encodingdefault}{\sfdefault}{bx}{n}

\def\gA{{\mathcal{A}}}

\def\gH{{\mathcal{H}}}

\def\gJ{{\mathcal{J}}}

\def\gL{{\mathcal{L}}}

\def\gN{{\mathcal{N}}}

\def\gP{{\mathcal{P}}}

\def\gR{{\mathcal{R}}}

\def\gT{{\mathcal{T}}}
\def\gU{{\mathcal{U}}}
\def\gV{{\mathcal{V}}}

\usepackage{stmaryrd}
\usepackage{amsfonts}
\usepackage{amssymb}
\usepackage{amsmath}
\usepackage{bbm}
\usepackage{xspace}
\usepackage{todonotes}
\usepackage{tabularx,booktabs,multirow}  
\usepackage{subcaption}
\usepackage{enumerate}
\usepackage{siunitx}

\newcommand{\ourattack}{\textbf{SLImE}\xspace}

\theoremstyle{plain}
\newtheorem{theorem}{Theorem}[section]

\theoremstyle{definition}
\newtheorem{definition}[theorem]{Definition}

\theoremstyle{remark}

\icmltitlerunning{Semantic Leakage from Image Embeddings}

\begin{document}

\twocolumn[
  \icmltitle{Semantic Leakage from Image Embeddings}



  \icmlsetsymbol{equal}{*}

  \begin{icmlauthorlist}
    \icmlauthor{Yiyi Chen}{yyy}
    \icmlauthor{Qiongkai Xu}{mac}
    \icmlauthor{Desmond Elliott}{dk}
    \icmlauthor{Qiongxiu Li}{yya}
    \icmlauthor{Johannes Bjerva}{yyy}
  \end{icmlauthorlist}

  \icmlaffiliation{yyy}{Department of Computer Science, Aalborg University, Copenhagen, Denmark}
  \icmlaffiliation{mac}{School of Computing, Macquarie University, Sydney, Australia}
  \icmlaffiliation{dk}{Department of Computer Science, University of Copenhagen, Copenhagen, Denmark}
  \icmlaffiliation{yya}{Department of Electronic Systems, Aalborg University, Copenhagen, Denmark}

  \icmlcorrespondingauthor{Qiongkai Xu}{qiongkai.xu@mq.edu.au}

  \icmlkeywords{Machine Learning, ICML}
  \vskip 0.3in
]



\printAffiliationsAndNotice{}  

\begin{abstract}
Image embeddings are generally assumed to pose limited privacy risk.
We challenge this assumption by formalizing \textbf{semantic leakage} as the ability to recover semantic structures from compressed image embeddings. 
Surprisingly, we show that semantic leakage does not require exact reconstruction of the original image.
Preserving local semantic neighborhoods under embedding alignment is sufficient to expose the intrinsic vulnerability of image embeddings.
Crucially, this preserved neighborhood structure allows semantic information to propagate through a sequence of lossy mappings. 
Based on this conjecture, we propose \textbf{S}emantic \textbf{L}eakage from \textbf{Im}age \textbf{E}mbeddings (\ourattack),
a lightweight inference framework that \textit{reveals semantic information from standalone compressed image embeddings}, incorporating a locally trained semantic retriever with off-the-shelf 
models, without training task-specific decoders.
We thoroughly validate each step of the framework empirically, from aligned embeddings to retrieved tags, symbolic representations, and grammatical and coherent descriptions.
We evaluate \ourattack across a range of open and closed embedding models, including \textsc{Gemini}, \textsc{Cohere}, \textsc{Nomic}, and \textsc{CLIP}, and demonstrate consistent recovery of semantic information across diverse inference tasks.
Our results reveal a fundamental vulnerability in image embeddings, whereby the preservation of semantic neighborhoods under alignment enables semantic leakage, highlighting challenges for privacy preservation.\footnote{We open-source all the code and datasets, facilitating further research.}
\end{abstract}


\section{Introduction}
Image embeddings have become a foundational component and widely traded commodity in agentic AI ecosystems, as the popularity of multimodal language models, such as Gemini~\citep{team2023gemini,comanici2025gemini} and ChatGPT~\citep{openai2023chatgpt}, has increased.
These embeddings are \textit{fixed-length representations that map high-dimensional visual inputs to compact vector spaces}. 
Frequently retrieved and shared, e.g., via Application Programming Interfaces (APIs), image embeddings are commonly assumed to preserve sensitive or private visual content due to their lossy and highly compressed nature~\citep{li2023learning}, which is reflected in the widespread deployment of embedding-based API services, such as Picone\footnote{\url{https://www.pinecone.io/}} and Weaviate\footnote{\url{https://weaviate.io/}}.

In this paper, we challenge this assumption and examine the privacy implications of sharing image embeddings under a realistic threat model that assumes no access to model internals and operates solely on image embeddings shared by potential victims.
We propose \textbf{S}emantic \textbf{L}eakage from \textbf{I}mage \textbf{E}mbeddings (\ourattack), a lightweight framework that challenges the assumption that compressed image embeddings are privacy-preserving. 


We show that \textit{the very objective for which image embeddings are optimized -- semantic retrieval -- can be utilized to undermine privacy}. 
Image embeddings are explicitly trained to preserve semantic similarity, enabling related images to be effectively retrieved and aligned with text, although they vary in architecture, training data, objectives and deployment settings~\citep{radford2021learning,jia2021scaling,xu2023bridgetower}. 
We demonstrate that such a retrieval-oriented property enables \textbf{semantic leakage}. 
By training a lightweight local retriever, an adversary can exploit the shared embedding space, where multimodal representations co-reside, to infer private and structured semantic information directly from image embeddings, using retrieval itself as the inference mechanism.

Building on the retrieved semantic signal, \ourattack i) leverages powerful off-the-shelf large language models (LLMs) to generate coherent sentences, simulating caption reconstruction attacks; ii) following an intermediate recovery of low-fidelity images from embeddings alone with an off-the-shelf diffusion model, then uses Vision-Language Models (VLMs) to perform adaptive attacks, progressively from identifying objects, relations, and further scene graphs. 
Importantly, this process does not rely on pixel-level reconstruction, access to the victim encoder, or task-specific decoders, but instead exploits the ability of contemporary generative models to amplify partial semantic cues into rich and coherent representations.

Finally, we show that \textit{semantic leakage persists even after multiple steps of lossy mappings and transformations}.
These transformations are induced by the local retriever, one-step alignment of target (victim) embeddings and subsequent LLM and VLM-based inferences.
Despite this cumulative loss, the local semantic neighbourhood remains sufficiently informative to induce leakage (cf. Section~\ref{semantic_neighborhood_preservation}).
This reveals a more fundamental privacy risk: leakage arises from the preservation of semantic meaning encoded in the representation, rather than from pixel-level detail or decoder-access.
As a result, privacy risks persist even for highly compressed embeddings, highlighting that privacy in multimodal systems must be considered at the level of semantic content rather than low-level vision features such as pixels.

Our contributions are as follows:
\begin{itemize}
    \item We formalize \textbf{semantic leakage} from the compressed image embeddings, and develop theoretical formulations of \textbf{semantic preservation via alignment} (cf. Section~\ref{sec:semantic_preservation}) to underpin this phenomenon and empirically evaluate it (cf. Section~\ref{sec:result_emprical}).
    \item We introduce \ourattack, a lightweight and model-agnostic framework for semantic inference from standalone image embeddings, exposing privacy risks without requiring access to original images, victim encoders, or task-specific decoders (Section~\ref{sec:framework}).
    \item We demonstrate the recoverability of image captions from image embeddings and conduct a series of adaptive semantic inference attacks that compounds privacy risks, across multiple embedding models, LLMs, and VLMs, as well as across diverse domains. The results reveal the ubiquitous challenges of semantic leakage in modern multimodal systems (Section~\ref{sec:semantic_attack},~\ref{sec:adaptive_attacks}).
\end{itemize}

While our approach leverages standardized components such as linear embedding alignment, contrastive retrieval and off-the-shelf LLMs/VLMs, our contribution does not lie in proposing new mechanisms, rather we introduce \textit{a conceptual and methodological shift in how semantic leakage from image embeddings is defined and evaluated}. 
We show that semantic leakage arises from the preservation of local semantic neighborhoods rather than exact ranking or reconstruction fidelity, formalize this notion and validate it empirically using a dual-reference evaluation protocol that disentangles embedding-induced semantic preservation from distortions introduced by discretization into tags and caption generation variations, rather than by embedding alignment itself.

\begin{figure*}[t!]
    \centering
    \includegraphics[width=0.85\linewidth]{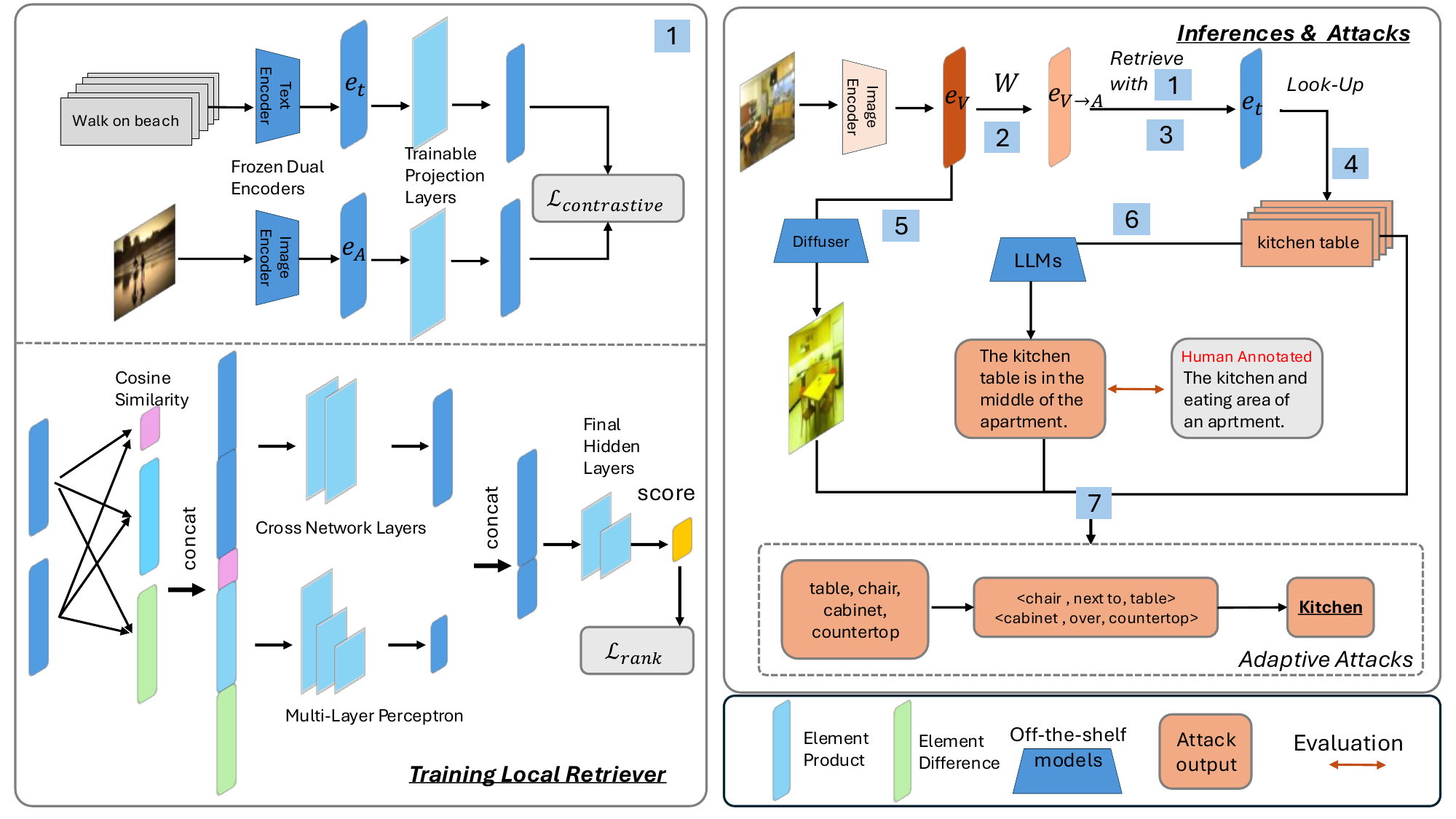}
        \caption{An overview of \ourattack Framework, consisting of (left) training a local retriever, and (right) inferences and attacks with the help of off-the-shelf models (see details in Section~\ref{sec:framework} and data example in Appendix~\ref{app:sample}).}
    \label{fig:framework}
\end{figure*}

\section{Related Work}
\textbf{Contrastive Learning and Image Embeddings} 
Contrastive learning has been widely adopted in learning image and vision-language representations.
The most well-known examples are CLIP, which trains a language and vision encoder using 400M paired examples~\citep{radford2021learning}, and SigLIP~\citep{zhai2023sigmoid}, a more compute-efficient version based on a sigmoid loss.
\citet{Chen2020ASF} uses contrastive learning to generate image embeddings, improving image classification accuracy. We use the CLIP model in some of our experiments that use open-weights models.
A key advantage of dual-encoder models is that they produce fixed-dimensional embedding models that can be precomputed, indexed, and reused across a wide range of tasks, making them particularly useful for large-scale retrieval systems and embedding-based APIs~\citep{Parekh2020CrisscrossedCE, Miech2021ThinkingFA,Hnig2023BiEncoderCF}.
We examine how this retrieval-enabling property can also induce security vulnerabilities.


\textbf{Semantic Leakage from Embeddings}
Prior work has largely focused on textual embeddings, where a fixed-size vector represents texts of varying lengths, to expose their semantic leakage.
The ability of such works ranges from recovering over half of the input words without preserving order~\citep{10.1145/3372297.3417270}, generating coherent and contextually similar sentences~\citep{li-etal-2023-sentence}, to exact reconstruction~\citep{morris2023text,chen2024text, chen2024typ}.
However, they mostly rely on training an expensive generator requiring a large amount of embeddings-text pairs in the target embedding space.
ALGEN uses one-step scaling to optimize alignment from the victim to the target embedding space and leverages a decoder trained on the attack space to reconstruct text from the aligned victim embeddings~\citep{chen-etal-2025-algen}. 
We use this few-shot alignment method to align image embeddings from the target space to the attack space for further analysis (cf. Section~\ref{sec:algen}).

While work exists in analyzing information leakage in encrypted images~\citep{Messadi2025ImageSE}, fewer have expolored the semantic leakage from image embeddings.
A line of work in this space reconstructs faces from supposed privacy-preserving face embeddings and launches attacks on facial recognition systems~\citet{Shahreza_2025_CVPR,11299120,Wang2025DiffMIBF}. 
Also leveraging reconstruction to leak crucial information,~\citet{Chen2025LeakyCLIPET} trains stable diffusion models to reconstruct semantically accurate images from CLIP embeddings.
In contrast, our work focuses on extracting semantic information from image embeddings without image reconstruction or training a massive decoder, with one-step alignment, a locally trained retriever, and off-the-shelf LLMs.

\textbf{Multimodal Language Models (MLLM)}
Recent surveys further highlight that the effectiveness of MLLMs largely stems from data-centric pretraining and adaptation strategies, as well as instruction tuning that leverages the reasoning priors of pretrained LLMs~\citep{Bai2024ASO}. 
At the same time, tuning and analysis studies reveal that while MLLMs are highly versatile, their performance and reliability depend critically on how visual representations are integrated and used during inference~\citep{Dang2024ExplainableAI, Huang2025KeepingYI}.
Recent work has demonstrated how to transform an LLM into a MLLM by training a simple linear projection from a vision encoder embedding space into an LLM input embedding space~\citep{tsimpoukelli2021multimodal,merullo2023linearly,liu2023visual}.
These models can be further improved through multi-stage fine-tuning on visual instruction tasks, or used directly for standard tasks, e.g. image captioning or visual-question answering.
Socratic Models embrace the zero-shot capabilities of pretrained models by prompt engineering guided multimodal tasks between the independent models to perform joint inference on a task-specific output~\citep{zeng2022socratic}.
For example, the objects, people, and places recognized by image classification systems can be transformed into a coherent sentence using an LLM.
In this work, we leverage these capabilities of MLLMs to expose and amplify the security vulnerabilities of image embeddings.

\section{\ourattack}\label{sec:framework}
We introduce a framework for studying \textbf{semantic leakage from image embeddings}, leveraging alignment and inference using a locally trained retriever and off-the-shelf models.
As illustrated in Fig.~\ref{fig:framework}, \ourattack consists of two stages: training (left) and inference (right).

Let $D$ denote a public dataset of images and their corresponding captions. 
For each image $i \in D$, we first process the caption(s) $\{C_i\}$ into a set of semantic tags $\{t_i\}$.
\numstep{1} We then encode the tags and images using pre-trained text and image encoders, yielding text embeddings \(\ve_t\) and image embeddings \(\ve_A\), respectively. 
Using these representations, we train a local retriever (cf. Section~\ref{sec:retriever}) -- based on a contrastive alignment model and a ranking module.
We assume access to a set of image embeddings produced by a victim visual encoder in an embedding space $\gV$. 
To enable inference, we align this victim space to the attack space $\gA$ using a learned linear alignment matrix \(\mW\) (\numstep{2}, cf. Section~\ref{sec:algen}). 
Given any victim image embedding \(\ve_V\), we obtain its aligned representation \(\ve_{V \rightarrow A}\) and use the local retriever to identify relevant text embeddings \(\ve_t\) \numstep{3} and their corresponding semantic tags \numstep{4} .

Building on this retrieved semantic signal, we conduct multi-stage inference attacks using off-the-shelf generative models:
i) \numstep{6} \textit{Text Reconstruction:}
For each aligned embedding \(\ve_{V \rightarrow A}\), we retrieve relevant tags using the local retriever and provide them as input to off-the-shelf LLMs to generate coherent textual descriptions, which are then evaluated against the original image captions (Section~\ref{sec:semantic_attack}).
ii) \textit{Adaptive Attacks:}
\numstep{5} We further employ an off-the-shelf diffusion model to generate low-fidelity images directly from \(\ve_{V \rightarrow A}\). \numstep{7}These generated images, together with the retrieved tags and text from the first stage, are fed into VLMs to identify \textit{objects, relations, and scene graphs}. This enables the incremental exploitation of leaked information across multiple stages of semantic inference (Section~\ref{sec:adaptive_attacks}).

\subsection{Local Retriever}\label{sec:retriever}
\paragraph{Preprocessing Tags}\label{preprocessing_tags}
Image captions describe visual content by referencing objects, entities, their relations, and broader scene context. 
A single image may therefore be associated with multiple objects, relations, and scene-level attributes expressed within a caption.
To analyze semantic leakage, we operate on semantic tags rather than full captions, which strips away word order and syntactic structure while retaining the core meaning-bearing units relevant for semantic inference~\citep{hendricks-nematzadeh-2021-probing}.
Concretely, we extract semantic tags from captions, including relational triples of the form \textit{$<$subject, verb, object$>$} and attribute tuples of the form \textit{$<$modifier, noun$>$}, using Stanza~\citep{qi2020stanza}.
The tags provide a retrieval-friendly yet compact representation of image semantics.
We have also experimented with training a retriever with tokens, which is sub-par compared to relational tags, consistent with prior work~\citep{Jiang2025T2IR1RI}.

\paragraph{Local Retriever}

As illustrated in Fig.~\ref{fig:framework}\numstep{1}, we first train the transformation layers from the pre-trained dual encoder to further align the image embeddings with semantically similar text embeddings with a contrastive learning objective.
Let $\{\ve_i\}^B_{i=1}$ be the image embeddings and $\{\ve_{t_j}\}^{N}_{j=1}$ be the semantic tag sets, with a binary matrix $\mM\in \{0,1\}^{B\times N}$ indicating which tags belong to which images. 
Using a learned temperature $\alpha$, we define similarities $s_{ij}=\alpha \ve_i^{\top} \ve_{t_j}$.
The image-to-tag objective is 
\[
\gL_{i\rightarrow t} = - \frac{1}{B} \sum^{B}_{i=1} \frac{1}{|\mathcal{P}_i|}
\sum_{j: \mM_{ij}=1} \log \frac{\exp(s_{ij})}{\sum_{k=1}^{N} \exp(s_{ik})},
\]

with $\gP_i=\{j:\mM_{ij}=1\}$.
Symmetrically, the tag-to-image objective is 
\[
\gL_{t\rightarrow i} = - \frac{1}{N} \sum^{N}_{i=1} \frac{1}{|\mathcal{J}_j|}
\sum_{i: \mM_{ij}=1} \log \frac{\exp(s_{ij})}{\sum_{k=1}^{B} \exp(s_{kj})},
\]
with $\gJ_j=\{i:\mM_{ij}=1\}$.
The total loss is 
\[
\gL_{\text{contrastive}}= \frac{1}{2}(\gL_{i\rightarrow t} +\gL_{t\rightarrow i}), 
\]
approximated by restricting the denominators to all positives and a subset of hard negatives per image.

After fine-tuning the pre-trained dual encoder with our data in the embeddings space, we further train a ranker on the embeddings with DCN v2 architecture~\citep{wang2021dcn}, consisting of cross-network and MLP modules, which takes the interaction features between normalized image and tag embeddings as follows:
\[
\phi(\ve_i, \ve_t)= [\ve_i; \ve_t; c;\vp;\vd ]\in \mathbb{R}^{4n+1}, 
\]
$\text{where } \vp=\ve_i \odot \ve_t \in \mathbb{R}^{n}, \vd=\ve_i -\ve_t \in \mathbb{R}^{n}$, $c$ is their cosine similarity, and $n$ is the embedding dimension.

The ranker is optimized using a grouped pairwise margin ranking loss with hard negative mining: for each image, positive tags are trained to rank higher than the highest-scoring negative tags by a fixed margin, and violations are penalized with a loss. 
The objective directly optimizes relative ordering within each image's candidate set, focuses learning on difficult confusions that affect ranking metrics, and avoids spurious cross-image comparisons, as follows:

\[
\gL_{\text{rank}} = \frac{1}{|\gU|} \sum_{u\in \gU } \frac{1}{|\gP_{u}| |\gH_{u}|}\sum_{p\in \gP_u} \sum_{n\in \gH_u} [m-(s_{u,p} - s_{u,n})]_+,
\]
where $[x]_+ := \max(x, 0)$,
$s_{u,p}$ and $s_{u,n}$ are the scores of positive and hard-negative tags for images $u$, $\gP_u$ and $\gH_u$ denote the positive and sampled hard-negative sets of ratio $\rho$, and $m$ is the margin, and $\gH_u = \operatorname{TopK}_{n\in \gN_u}(s_{u,n}),\quad |\gH_u|=\max(1,\lfloor\rho|\gN_u|\rfloor)$.
At inference, for each image $u$, we score all candidate tags $t\in \gT$ with the trained DCN ranker $s(u,t)$ and return the top-$K$ tags as the local retriever output. \

\subsection{Embedding Space Alignment}\label{sec:algen}


To reduce the discrepancy between the attack and the victim image embedding spaces, ${\ve_{V}}$ is mapped to the space $A$ with an learned linear mapping matrix $\mW$, i.e., $\ve_{V\rightarrow A} = \ve_{V}\mW$.
We obtain the optimal alignment matrix $\mW$ is by solving the following least-squares minimization:
\begin{equation}
    \min_{\mW}
    \|\mE_{A}- \mE_{V} \cdot \mW\|^{2},
\end{equation}
where $\mE_{V}=[\ve_{V}^{1\top}, \cdots, \ve_{V}^{b\top}]^{\top} \in \mathbb{R}^{b\times m}$ is the victim embedding matrix, and $\mE_{A}=[\ve_{A}^{1\top}, \cdots, \ve_{A}^{b\top}]^{\top} \in \mathbb{R}^{b\times n}$ is the attack embedding matrix, and $b$ is the number of alignment samples, whereas $m$ and $n$ are the embedding dimensions in victim and attack spaces respectively.
The solution to this least squares loss is:
\begin{equation}\label{eq:w}
\mW = (\mE^{T}_V \mE_{V})^{-1} \mE_{V}^{T} \mE_{A},
\end{equation}
where $(\mE^{T}_V \mE_V)^{-1} \mE_{V}^{T}$ is the \textit{Moore-Penrose Inverse} of $\mE_V$ (see the detailed derivation in Appendix~\ref{normal_equation}).
The aligned embedding matrix from $V$ to $A$ is thus: 
\begin{equation}\label{eq:aligned}
\mE_{V \rightarrow A} = \mE_{V}  \mW,
\end{equation}
where $\mE_{V \rightarrow A}\in \mathbb{R}^{b\times n} $. Implementing this alignment does not require any training; it is a one-step linear scaling. 

\subsection{Semantic Preservation via Alignment}\label{sec:semantic_preservation}


\ourattack operates in a standard retrieval-based setting, where image and text embeddings are compared via cosine similarity after linear alignment, using ALGEN (cf. Section~\ref{sec:algen}).
For our experiment, all image and text embeddings are $l_2$-normalized, the alignment matrix $W$ is learned using ~\eqref{eq:w}, and retrieval and ranking are performed by inner product (cosine similarity under normalization).



\begin{definition}[Semantic Neighborhood]\label{def:semantic_neighborhood}
Let $\gT$ be the tag vocabulary with embeddings $\{\ve_t \in \mathbb{R}^{d}\}_{t\in\gT}$.
For any tag $t\in\gT$, its semantic neighborhood of size $m$ is defined as
\[
\gN_m(t) := \operatorname{Top}\text{-}m_{u\in\gT}
\ \langle \ve_t, \ve_u\rangle,
\]
i.e., the $m$ most similar tags under cosine similarity.
\end{definition}

\begin{definition}[Semantic Neighborhood Preservation]\label{semantic_neighborhood_preservation}
Let $G_i := \mathrm{TopK}(\ve_{A,i})$ be the reference tag set for image $i$, and
$P_i := \mathrm{TopK}(\ve_{V,i}\mW)$ be the aligned retrieval result.
For each $g\in G_i$, let $\gN_m(g)$ denote its semantic neighborhood.
We say that semantic neighborhood preservation holds at scale $(m,K)$ if
the retrieved tags in $P_i$ lie within the semantic neighborhoods of $G_i$. 
\end{definition}

We quantify this property using neighborhood precision, recall, and their harmonic mean $F1_i(m,K)$ (cf. Section~\ref{sec:result_emprical}).




With \ourattack, we show that key semantic information can be recovered through retrieval even after lossy transformations, without requiring exact matching or a task-specific decoder.
In particular, semantic neighborhood preservation is sufficient to identify salient objects, attributes, and relations of an image, even when exact tag retrieval fails.
Our experiments validate that local neighborhood structure consistently provides enough semantic information to support inference and reconstruction, including adaptive attacks (cf.~Sections~\ref{sec:semantic_attack}, ~\ref{sec:adaptive_attacks}).

\section{Experimental Setup}


\subsection{Models}
We conduct attacks on both commercial and open-sourced embedding models, including \textsc{Gemini}, \textsc{Nomic}, \textsc{Clip}, and \textsc{Cohere}. 
To simulate attackers, we experiment with open-sourced and commercial LLMs, including \textsc{Gpt-5.2}, \textsc{Gemini-3-pro}, \textsc{Deepseek-v3.2},  \textsc{Qwen3-235B}, \textsc{Cohere-v2}, and \textsc{Minmax-m1}, and VLMs, including~\textsc{Gemini-flash}, \textsc{Gpt-5.1}, and \textsc{Qwen2.5-vl}.
We train our contrastive learning model and retrieval ranker on the backbone dual encoder from \textsc{Kandinsky-2.2}, which is a diffusion model uses pre-trained \textsc{clip} model as a text and image encoder. See Table~\ref{tab:model_detail} for full details.

\subsection{Datasets}
We follow the original COCO splits~\citep{chen2015microsoft}, using the full training set and subsampling 500 samples each from the validation and test sets for retriever training/validation and attack evaluation, respectively.
We use a subset of 500 samples of nocaps~\citep{agrawal2019nocaps}, consisting of 250 near-domain and 250 out-domain images compared to COCO, to explore the generalizability of \ourattack on cross-domain attacks.

\subsection{Evaluation Metrics}
We use Recall, Precision, and F1 scores to measure retrieval performances.
For embedding alignment, we use cosine similarity (COS) to measure the similarity of attack embeddings and aligned victim embeddings. 
We use string-matching metrics such as BLEU~\citep{papineni-etal-2002-bleu}, ROUGE (1/2/L)~\citep{lin-och-2004-automatic}, and METEOR~\citep{denkowski-lavie-2014-meteor} to evaluate LLM-generated texts against human-written texts. An image can be associated with multiple captions; we use the \textit{best match} strategy for robust and realistic attacker-view evaluation (cf. Appendix~\ref{eval_llm_text}).


\subsection{Semantic Information Propagation and Evaluation}\label{sec:propagation}
We view \ourattack as a propagation of semantic information through a sequence of lossy mappings,
\[
\ve_V
\;\xrightarrow[\numstep{2}]{\;\mW\;}\;
\ve_A
\;\xrightarrow[\numstep{3}]{\;\gR_K\;}\;
\{\ve_t\}
\;\xrightarrow[\numstep{4}]{\;\gT\;}\;
t
\;\xrightarrow[\numstep{6};\numstep{7}]{\;\text{L(V)LMs} \;}\;
\{C\}
\]
where each stage introduces a distinct gap. 

\numstep{2} The alignment maps victim embeddings into the attack space via the alignment matrix $\mW$, with cosine similarities used to measure alignment quality and reflect the gap (cf. Section~\ref{sec:semantic_attack}).
\numstep{3} A retrieval operator $\gR_K$ maps each aligned embedding to a Top-$K$ set of tag embeddings based on similarity, enabling evaluation of local semantic neighborhood preservation (cf.~Definition~\ref{semantic_neighborhood_preservation}, Section~\ref{sec:result_emprical}).
\numstep{4} The Look-Up stage maps retrieved tag embeddings to a vocabulary $\gT$ of relational tags, introducing discretization without additional geometric structure.
\numstep{6};\numstep{7} Finally, symbolic tags are mapped to natural-language captions using LLMs and VLMs, amplifying partial semantic information into coherent descriptions. Importantly, our analysis does not require exact semantic preservation across all stages; preserving local semantic neighborhoods at retrieval suffices to enable downstream reconstruction despite substantial distortion.



\section{Results and Analysis}
We denote the ground-truth relational tags $\{t_{\mathrm{gt}}\}$  and LLM-generated captions $\{C_{\mathrm{gt}}\}$, and the human annotated captions $\{C_{h}\}$ , respectively.
Tags and captions obtained directly from the attack embedding $\ve_A$ are denoted by $\{t_A\}$ and $\{C_A\}$.
Correspondingly, tags and captions obtained from aligned victim embeddings $\ve_V\mW$ are denoted by $\{t_{V\!\rightarrow\!A}\}$ and $\{C_{V\!\rightarrow\!A}\}$.

\subsection{Semantic Neighborhood Preservation via Alignment}\label{sec:result_emprical}

We empirically validate Definition~\ref{semantic_neighborhood_preservation} by operationalizing it into its constituent conditions and evaluating each component directly.

\textbf{Reference Semantic Signal.} For each test image $i$, we define the reference tag set $G_i=\mathrm{TopK}(\ve_{A,i})$ corresponding to the Top-$K$ tags under the attack emebeddings. This allows us to assess whether aligned victim embeddings preserve the same local semantic structure as the attack embeddings, without relying on external annotations, e.g., $\{t_{gt}\}$.

\textbf{Aligned Retrieval.} Given aligned victim embeddings $\ve_{V,i}\mW$, we obtain the retrieval set $P_i := \mathrm{TopK}(\ve_{V,i}\mW)$, following Definition~\ref{semantic_neighborhood_preservation}.

\textbf{Cohort Coverage.} For a reference tag $t\in G_i$, we count a hit if at least one retrieved tag lies in its semantic cohort, $\mathrm{hit}_i(t)
:= \mathbf{1}\!\left[\exists\, u\in P_i \text{ s.t. } u\in \mathcal N_m(t)\right]$, and define neighborhood recall as $\mathrm{Recall}_i := \frac{1}{|G_i|} \sum_{t\in G_i} \mathrm{hit}_i(t)$, which directly estimates the fraction of reference cohorts covered by aligned Top-$K$ retrieval.

\textbf{Non-Triviality.} To ensure that cohort coverage is not achieved trivially by retrieving semantically unrelated tags, we measure neighborhood precision. 
A retrieved tag $u\in P_i$ is considered explained if it belongs to the semantic cohort of any reference tag: $\text{expl}_i:= \mathbf{1}\!\left[\exists\, t\in G_i \text{ s.t. } u\in \mathcal N_m(t)\right]$, with precision defined as $\mathrm{Precision}_i := \frac{1}{|P_i|} \sum_{u\in P_i}\text{expl}_i(u) $.

We present the trade-off between coverage and specificity using the $F1$ score. 

\begin{figure}
    \centering
    
    \includegraphics[width=0.9\linewidth]{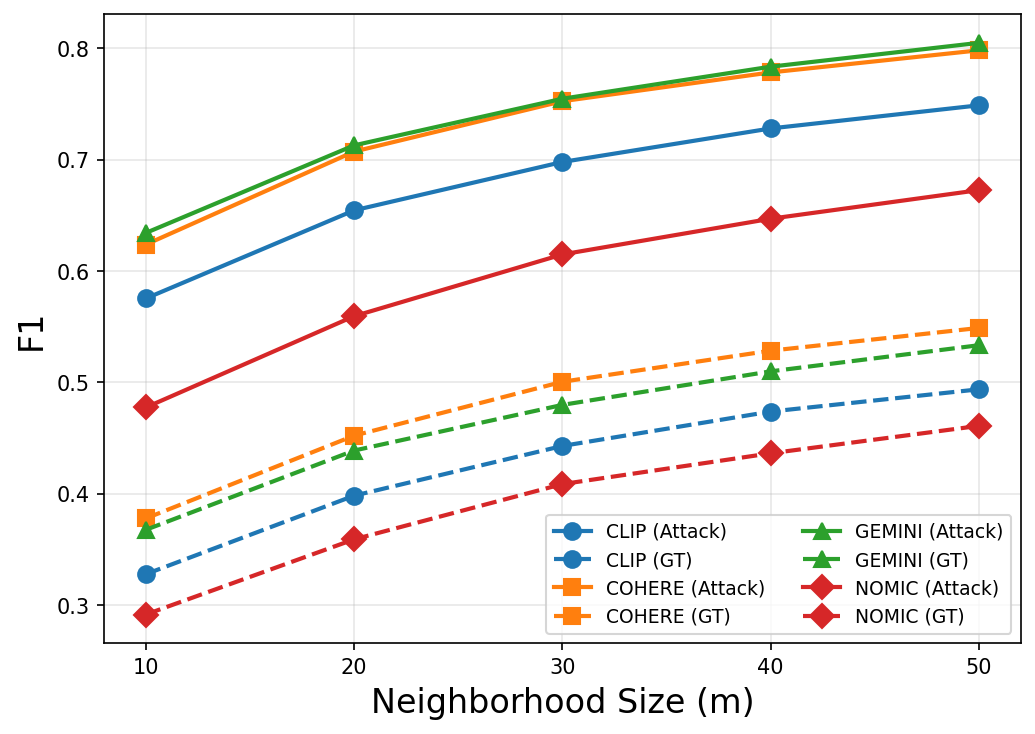}
    \caption{Semantic Neighborhood Preservation under Alignment. F1 scores across neighborhood size $m$ for aligned Top-$K=30$ retrieval using $10,000$ alignment samples, across victim embeddings, evaluated against \(\{t_{gt}\}\) (dotted lines) and \(\{t_{A}\}\)(solid lines).}
    \label{fig:f1_m_size}
\end{figure}

\textbf{Effect of Neighborhood Size $m$ on Semantic Neighborhood Preservation }
As shown in Fig.~\ref{fig:exact_tag_retrieval}, exact tag retrieval remains consistently low (below 0.25) across $K \in \{5,10,15,20,25,30\}$ and alignment sample sizes $\{1,10,100,1000,10000\}$, whether evaluated against $\{t_{gt}\}$ or $\{t_A\}$.
Nevertheless, retrieval performance increases monotonically with both the number of alignment samples and $K$.
In comparison, fixing retrieval $K=30$ with $10,000$ samples to learn $\mW$, we report $F1(m,K)$ in Fig.~\ref{fig:f1_m_size}. 
Across all victim embedding models, F1 increases monotonically with $m$ and saturates at $m=50$, and F1 scores above 0.5 mostly are not an artefact of trivial matching, especially when evaluated against tags retrieved for attack embeddings $\{t_{A}\}$, reaching $0.8$.
This trend indicates that relatively small semantic neighborhoods already suffice to achieve a stable cohort coverage under alignment, consistent with Definition~\ref{semantic_neighborhood_preservation}. 
The steep initial increase in F1 indicates that aligned retrieval reliably recovers semantically related tags without requiring exact rank or identity preservation.
The subsequent saturation suggests that leakage is driven by local neighborhood preservation rather than global semantic overlap, with most informative tags retrieved early.
By contrast, F1 scores computed against ground-truth relational tags $\{t_{gt}\}$ remain substantially lower across $m$.

Nevertheless, aligned retrieval consistently preserves semantic neighborhoods, showing that this condition is sufficient, albeit not necessary, for semantic leakage to occur.
In particular, leakage does not require a strong retriever in the classical information-retrieval sense; it suffices that alignment preserves local semantic neighborhoods, a property explicitly encouraged by embedding models.

\subsection{Text Reconstruction from Semantic Tags}\label{sec:semantic_attack}
After retrieving sufficient relational tags for semantic preservation, we can use LLMs to generate text from the tags to explore whether grammatically correct and syntactically coherent information can be recovered.

As previously described, there can be a gap between $\{t_{gt}\}$ and $\{C_{h}\}$ (cf. Section~\ref{sec:propagation}).
We first use closed- and open-source LLMs to generate captions $\{t_{gt}\}$, i.e., $\{C_{gt}\}$, and evaluate against the human annotated descriptions of the images, i.e., $\{C_{h}\}$.
As shown in Table~\ref{tab:eval_gt_tag_llm}, \textsc{Deepseek} achieves Rouge-L comparable to the best-performing proprietary model \textsc{Gemini}, and the results are strong across metrics, recovering more than half of the information by \textsc{Gemini-3-Pro}, \textsc{Deepseek-v3.2}, and \textsc{Gpt-5.1}.
We therefore use \textsc{Deepseek-v3.2} for generating $\{C_{A\rightarrow V}\}$ from $\{t_{A\rightarrow V}\}$, benefiting from its open-source accessibility and cost efficiency while maintaining competitive caption quality.

\begin{table}[!t]
    \centering
     \caption{Evaluating LLM-Generated Captions $\{C_{gt}\}$ from Ground-Truth Relational Tags $\{t_{gt}\}$ against Huamn-Annotated Captions $\{C_{h}\}$.} 
     \label{tab:eval_gt_tag_llm}
  \begin{center}
      \begin{sc}
      \resizebox{0.9\linewidth}{!}{
    \begin{tabular}{c|ccccc}
    \toprule
        \textbf{LLM} & \textbf{BLEU-4} & \textbf{Rouge-1} & \textbf{Rouge-2} & \textbf{Rouge-L} & \textbf{Meteor} \\ 
        \midrule
        \textbf{Gemini-3-pro} & 25.39 & 66.41 & 40.63 & \textbf{60.53} & 62.76 \\ 
        \textbf{Deepseek v3.2} & 18.89 & 61.86 & 33.90 & \underline{54.36} & 56.52 \\ 
        \textbf{GPT 5.2} & 14.34 & 58.07 & 28.57 & 50.22 & 55.71 \\ 
        \textbf{Qwen3-235b} & 13.88 & 56.80 & 27.80 & 48.55 & 53.18 \\ 
        \textbf{Cohere v2} & 11.78 & 54.78 & 25.24 & 45.86 & 51.85 \\ 
        \textbf{Minimax-M1} & 10.57 & 49.67 & 21.51 & 42.96 & 45.25 \\ 
        \bottomrule
    \end{tabular}}
    \end{sc}
    \end{center}
\end{table}

\begin{figure}[t]
    \centering
   
    \includegraphics[width=\columnwidth]{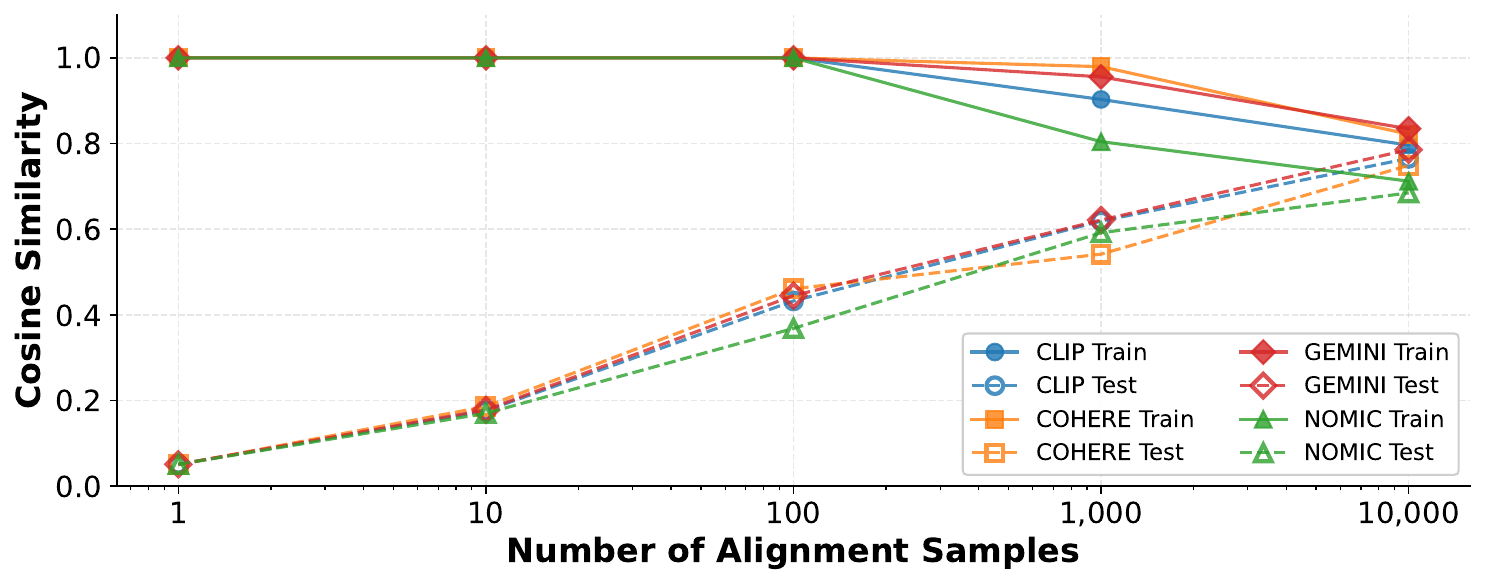}
\includegraphics[width=\columnwidth]{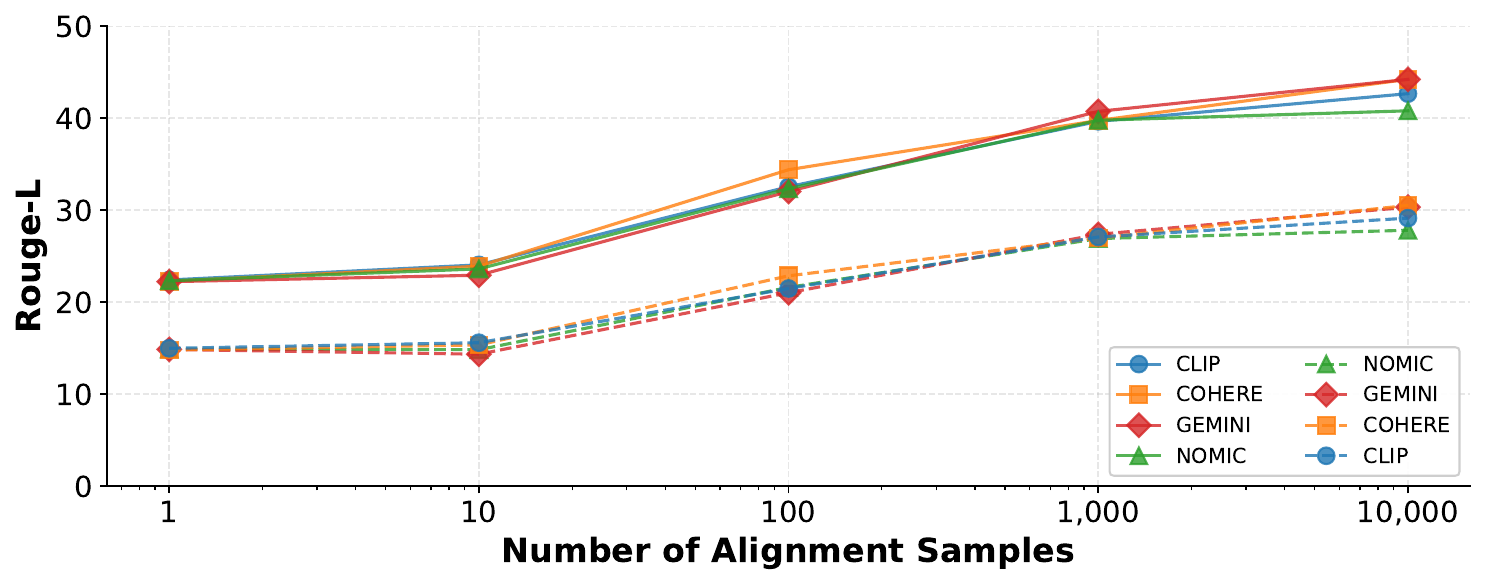}
 \caption{Semantic Leakage of Image Embeddings by Alignment Samples. (Top) Cosine Similarity between Attack and Aligned Embeddings; (Bottom) $\{C_{V\rightarrow A}\}$ evaluated against $\{C_{gt}\}$ (solid lines) and $\{C_{h}\}$ (dotted lines) against at $K=10$ in Rouge-L.}
    \label{fig:llm_cosine_rougel}
\end{figure}



Fig.~\ref{fig:llm_cosine_rougel} (top) shows that increasing the number of samples used to learn $\mW$ leads to progressively better alignment, with the mapped victim embeddings $\ve_{V\rightarrow A}$ moving closer to attack-space embeddings on the test set.
We further observe that \textsc{Gemini} embeddings achieve the strongest alignment among the four victim embedders, although the differences are minor.
To select an appropriate value of $K$ for subsequent experiments, considering 15 as the average number of tags $\{t_{gt}\}$ for each image (cf. Appendix~\ref{stats_app}), we perform an ablation study over $K \in \{5, 10, 15, 20, 25\}$.
For each setting, we generate captions from the retrieved tags and evaluate them against $\{C_{gt}\}$ using Rouge-L.
As shown in Fig.~\ref{fig:ablations_gemin_topk}, $K=10$ yields the best performance and is therefore used in subsequent evaluations, which further validate that the key information is retrieved early.
Across victim embedders, using $10\text{K}$ alignment samples recovers nearly half of the semantic information relative to $\{C_{gt}\}$.
Notably, even with a single alignment sample, the Rouge-L score already exceeds 20.
While performance is lower when evaluated against $\{C_h\}$, Rouge-L scores remain consistently in the range of 10 - 30 (see more results in Fig.~\ref{fig:results_full_text_reconstruction}).

Overall, our results show that image embeddings remain vulnerable to semantic leakage even when only \textit{a single data point} is leaked, and that this risk is largely embedder-independent.
We further validate that exact-match retrieval measures systematically underestimate leakage risk, as meaningful semantic information can already be inferred from local neighborhoods.
Together, these findings suggest that \textit{privacy in embedding-based systems must be evaluated at the level of semantic inference, rather than pixel reconstruction or exact retrieval accuracy}.

\subsection{Adaptive Attacks}\label{sec:adaptive_attacks}
We define \textbf{adaptive attacks} as multi-stage inference task in which the attacker incrementally conditions each query on semantic content inferred in earlier stages.
The attacker progressively infers increasingly structured semantic representations, such as objects, relations, and scene-level context, and reuses these intermediate inferences to guide subsequent queries.
This staged design models a realistic adversary who reasons over partial and noisy outputs, enabling us to isolate how semantic leakage compounds across inference steps.

Each stage operates on different combinations of previously retrieved tags $\{t_{V\rightarrow A}\}$, generated captions $\{C_{V\rightarrow A}\}$, and low-fidelity images inferred from $\ve_{V \rightarrow A}$ (an example in Fig.~\ref{fig:lofi_reconstruction}).
For evaluation, the objects, relations, and scene graphs are also identified using VLMs from $\{t_{gt}\}$, $\{C_{h}\}$, and original images.
We provide a complete illustration of the inputs and outputs for a test sample in Appendix~\ref{app:sample}.



\begin{figure}[t!]
    \centering
    \includegraphics[width=0.8\linewidth]{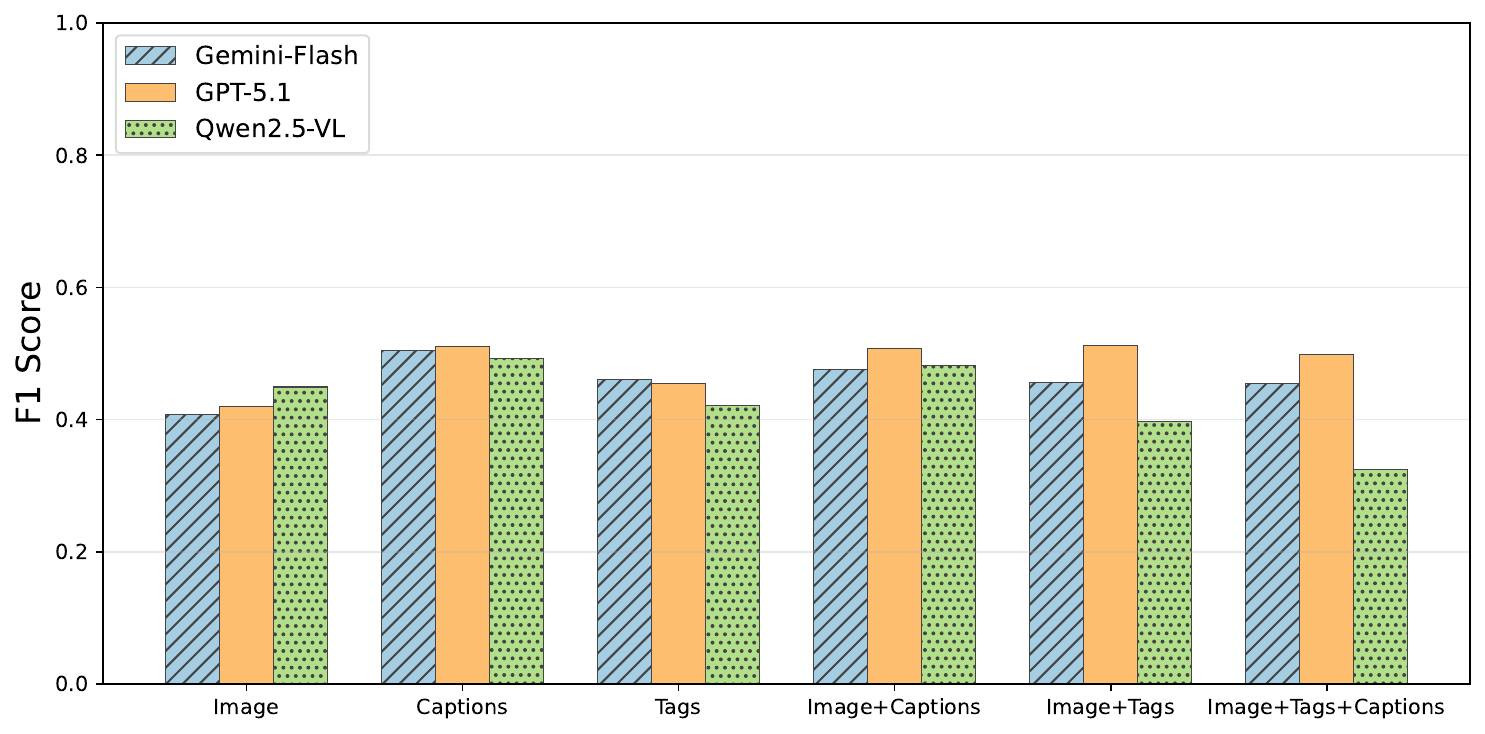}
        \caption{Identified Objects with VLMs across Settings in F1.}

    \label{fig:macro_f1_objects}
\end{figure}

\textbf{Bootstrapping Objects}
Using VLMs, we infer objects from low-fidelity images alone, from retrieved tags and generated captions, or from their combinations, shown in Fig.~\ref{fig:macro_f1_objects}.
Evaluation against objects identified from original inputs using the same VLM shows that low-fidelity images already recover nearly half of the objects.
Conditioning on retrieved tags or captions further improves performance, with \textsc{GPT-5.1} and \textsc{Gemini-Flash} achieving the strongest gains.

\begin{figure}[t]
    \centering

    \includegraphics[width=0.8\linewidth]{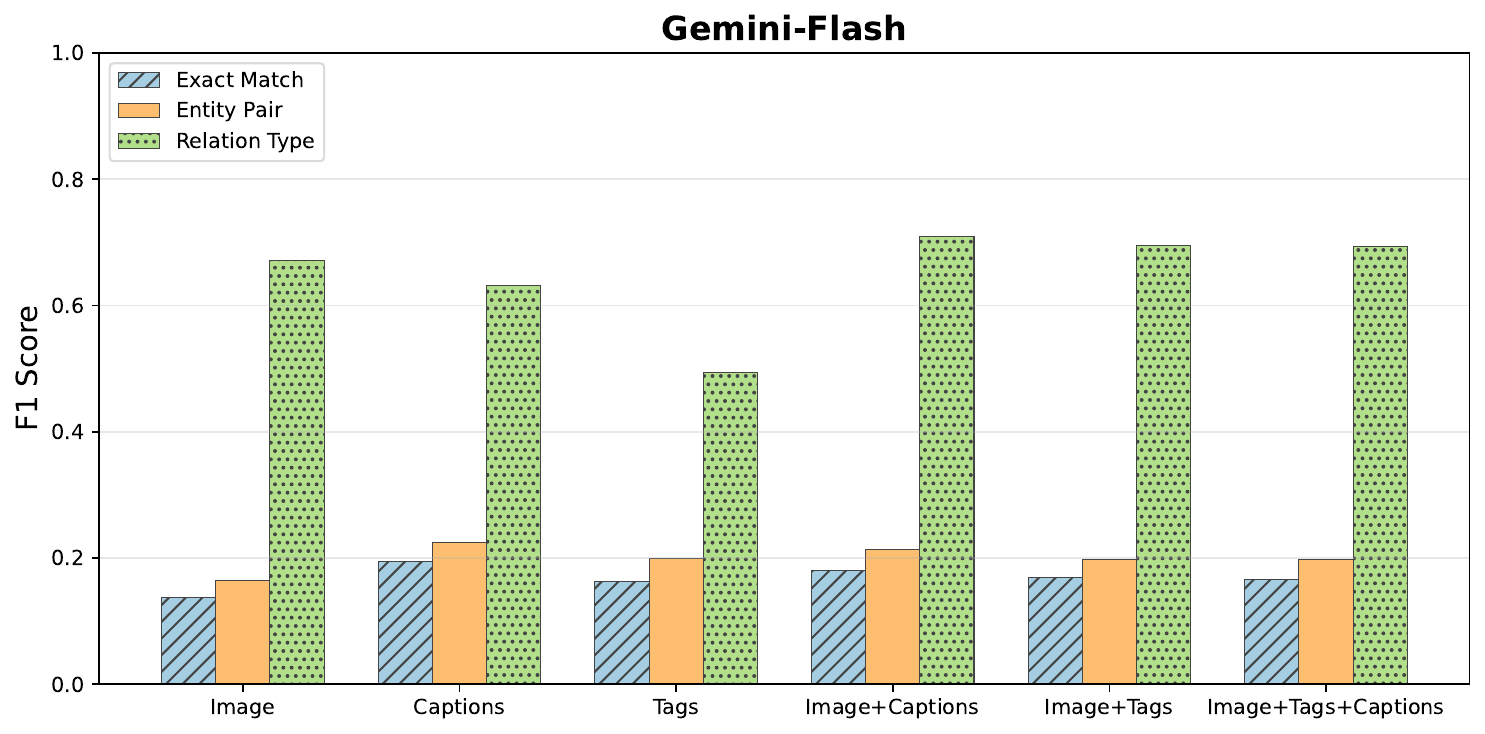}
    \caption{Identified Relations across Settings in F1.}

    \label{fig:relations_gemini_f1}
\end{figure}

\textbf{Inferring Relations}
We report F1 scores for exact matching of relation triples, entity pairs, and relation types, as shown in Fig.~\ref{fig:relations_gemini_f1} (see evaluations using \textsc{Gpt-5.1} and \textsc{Qwen2.5-VL} in \ Fig.~\ref{fig:qwenvl_gpt_scene_graphs}).
Across settings, relation types are recovered with high accuracy, reaching over 50\% and up to 70\% F1 when using \textsc{Gemini-Flash}, particularly when conditioning on images alone or in combination with captions and tags.
The set of relation types is defined in Appendix~\ref{relations_app}.

\begin{figure}
    \centering
    \includegraphics[width=0.8\linewidth]{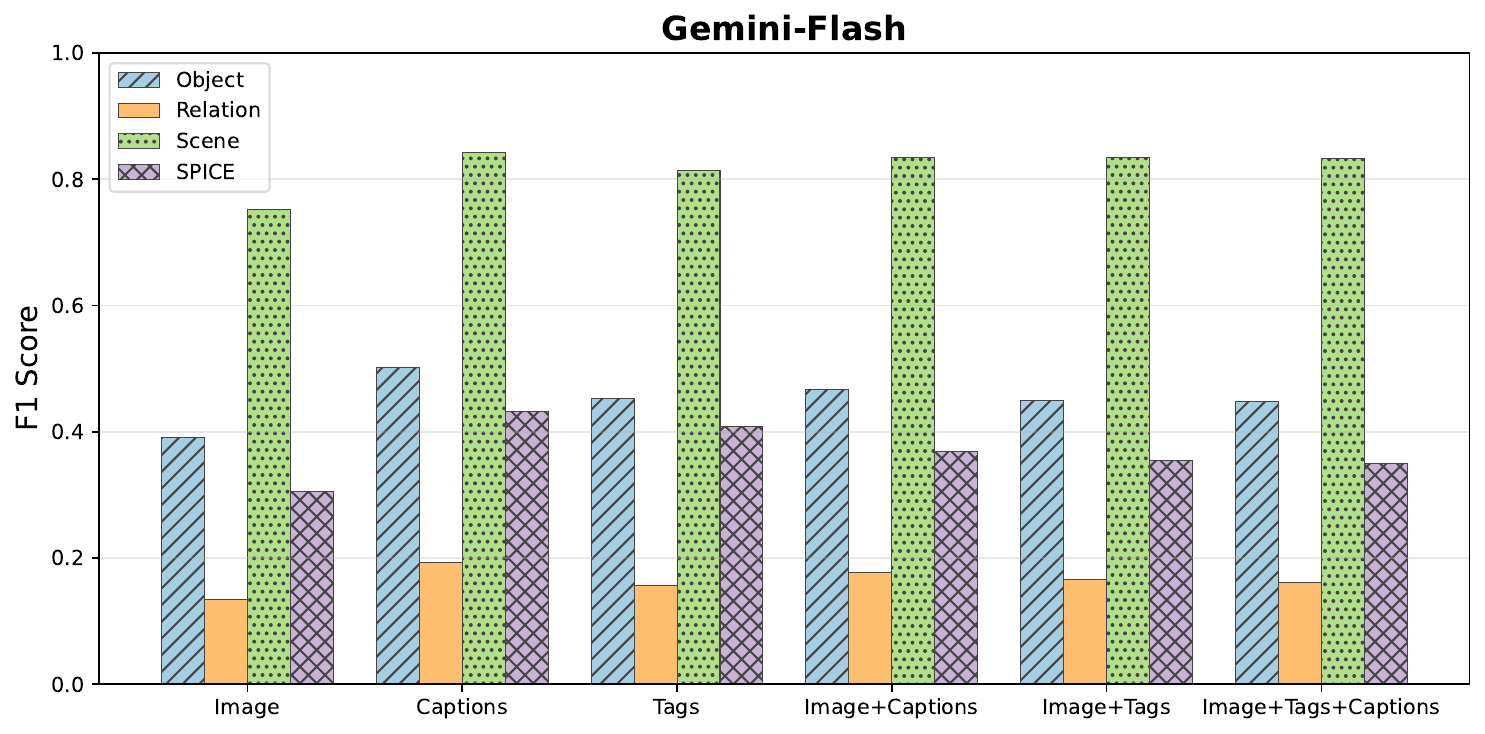}
        \caption{Identified Scene Graphs across Settings in F1.}

    \label{fig:scene_graphs_gemini_flash}
\end{figure}

\textbf{Inferring Scene-level Semantics}
Across all VLMs, we observe consistent recovery of object- and scene-level semantics under multimodal inference, with scene F1 remaining high ($\approx 0.75$–$0.88$) even when relation recovery is limited, as shown in Fig.~\ref{fig:scene_graphs_gemini_flash}.
Under our setting, object backfilling is disabled during relation and scene inference, prioritizing precision over recall.
Notably, retrieved tags alone already enable strong object and scene inference, indicating that retrieval is sufficient to induce semantic leakage.
High-level scene semantics can thus be reliably inferred even when fine-grained relations are sparsely recovered, for example, with relation F1 as low as 0–10\% for \textsc{GPT}- and \textsc{Qwen}-based VLMs, as in Fig.~\ref{fig:qwenvl_gpt_scene_graphs}.
Finally, combining modalities does not always improve performance, as low-fidelity images alone often provide sufficient semantic information.

\begin{figure}
    \centering
        \includegraphics[width=0.95\linewidth]
        {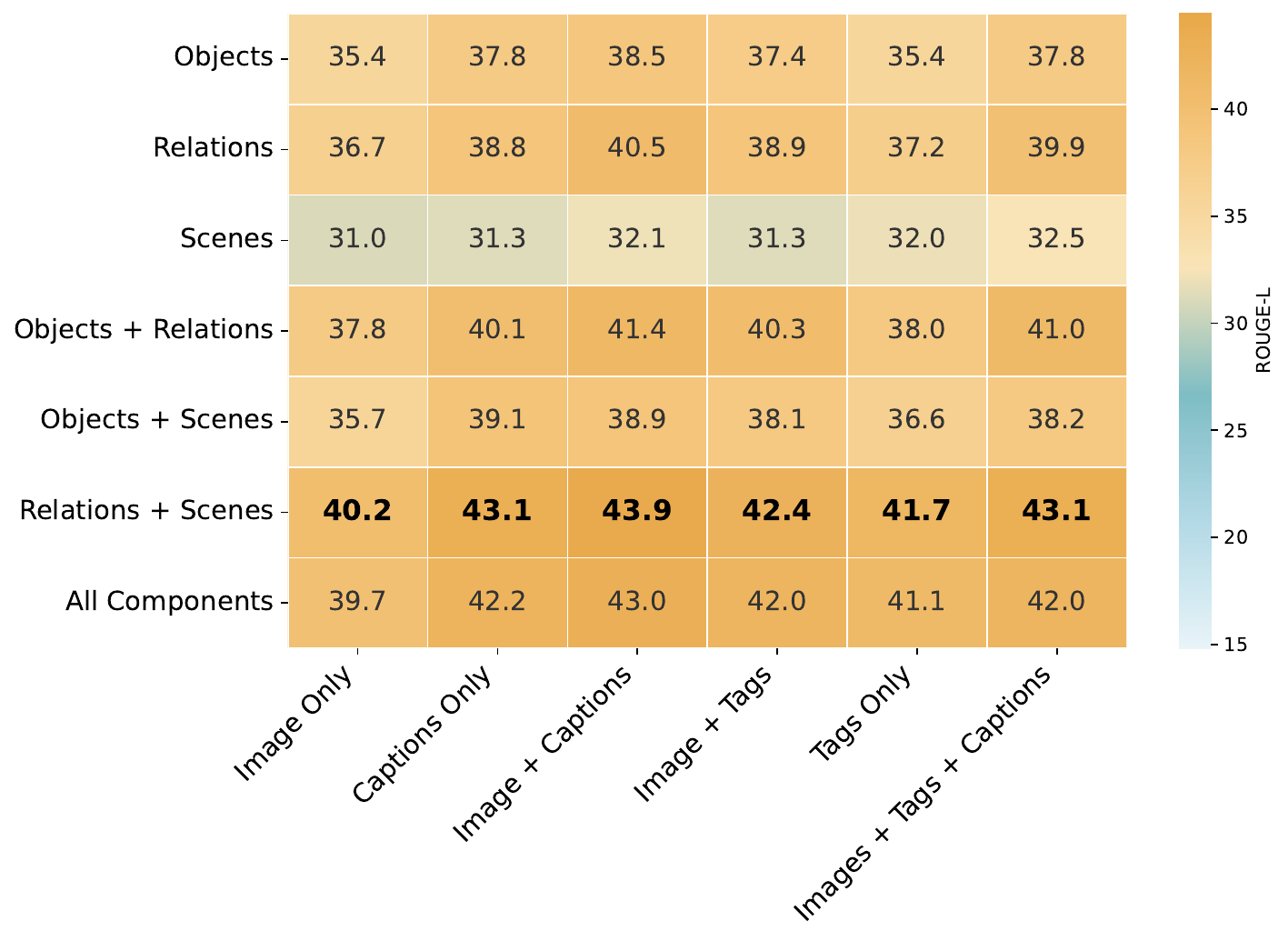}
        \caption{Evaluate Generated Captions \(\{C_{V \rightarrow A}\} \) against \(\{C_{gt}\}\) across settings, the best Rouge-L score  for each column is \textbf{bolded}.}
        \label{fig:heatmap_rougel_ablation_vlm_gt}
\end{figure}

\textbf{Ablation Studies}
Based on prior results showing that \textsc{Gemini-Flash} performs best, particularly for scene graph inference, we use it for the ablation study, that systematically varies which intermediate semantic representations (objects, relations, scenes) are provided to generate captions.
As shown in Fig.~\ref{fig:heatmap_rougel_ablation_vlm_gt}, conditioning on relations and scenes consistently recover captions more than other combinations (see Appendix~\ref{sec:ablation_app}). Relations and scenes capture how objects interact and provide global context; adding extra isolated objects can introduce redundant or even conflicting cues. Consequently, conditioning on relations and scenes alone often yields captions that are more coherent and informative than combining all semantic representations.

\begin{table}[t]
    \centering
    \caption{Cross-Domain Evaluation on  \(\{C_{V\rightarrow A}\}\) against \(\{C_{h}\}\). The best Rouge-L scores are \textbf{bolded}.}
      \resizebox{0.9\linewidth}{!}{
    \begin{tabular}{cccccc}
    
    \toprule
          \textbf{Victim Embedder} & \textbf{BLEU-4} & \textbf{ROUGE-1} & \textbf{ROUGE-2} & \textbf{ROUGE-L} & \textbf{METEOR} \\ 
        \midrule
         \textbf{Near-Domain} & & & & &  \\
        clip & 11.16 & 43.29 & 16.40 & 38.27 & 35.33 \\ 
        cohere & 12.45 & 44.03 & 17.39 & \textbf{38.69} & 36.66 \\ 
        gemini & 11.62 & 43.66 & 17.03 & 38.22 & 36.19 \\ 
      nomic & 9.59 & 39.88 & 14.16 & 34.85 & 32.18 \\ 
      \midrule
      \textbf{Out-Domain}& & & & & \\
        clip & 8.20 & 40.06 & 12.87 & 34.29 & 33.06 \\ 
        cohere & 9.36 & 41.92 & 14.13 & \textbf{36.07} & 35.18 \\ 
     gemini & 9.01 & 40.85 & 13.75 & 35.00 & 34.27 \\ 
    nomic & 7.44 & 37.42 & 11.20 & 32.14 & 30.59 \\ 
       \bottomrule
    \end{tabular}}
    \label{tab:cross_domain_reconstruction}
\end{table}

\subsection{Cross-domain Semantic Leakage}
We use \textsc{nocaps} for cross-domain evaluation.
Table~\ref{tab:cross_domain_reconstruction} shows that semantic leakage generalizes beyond the alignment domain, albeit with a moderate performance drop.
Across all victim embedders, moving from near-domain to out-of-domain evaluation leads to consistent but limited degradation across metrics, indicating that the inferred semantic structure remains partially transferable (see more results in Appendix.~\ref{cross-domian_appendix}).

\section{Discussion and Conclusion}
We systematically study semantic leakage in image embeddings through \ourattack.
We show that meaningful semantic content can be inferred from image embeddings alone via alignment, retrieval, and multi-stage inference, even under lossy transformations and domain shift.
Our experiments demonstrate that leakage arises from the preservation of local semantic neighborhoods, a core objective of retrieval-oriented embeddings.
These findings indicate that privacy risks in embedding-based systems cannot be mitigated solely by limiting reconstruction fidelity or restricting decoder access.
As long as embeddings are optimized for semantic retrieval, they inherently expose semantic information that can be exploited through inference.
This characterizes semantic leakage as an intrinsic vulnerability of image embeddings, rather than an artifact of specific models or attacks.
We overall offer a new perspective on semantic-level analysis of image embedding privacy and encourage future work on mitigation at this level.

\section*{Acknowledgements}
Y.C. and J.B. are funded by the Carlsberg Foundation, under the Semper Ardens: Accelerate programme (project no.~CF21-0454) and the Novo Nordisk Foundation under the Ascending Data Science Investigator programme (grant no.~NNF24OC0092972).
We further acknowledge the support of the AAU AI Cloud and express our gratitude to DeiC for providing computing resources on the LUMI cluster.
Y.C. expresses her gratitude to the Danish Ministry of Higher Education and Science for the EliteForsk Travel Grant.
Q.X. acknowledges support from 2024 FSE Strategic Startup and FSE Travel Grant.
D.E. was supported by the European Union’s Horizon 2020 research and innovation program under
grant agreement No. 101135671 (TrustLLM) and a research grant (VIL53122) from Villum Fonden.

\section*{Impact Statement}
This work investigates privacy risks arising from semantic leakage in image embedding models. The findings will increase the awareness of privacy risks at the semantic level, beyond pixel-level reconstruction or model inversion attacks. By framing semantic leakage as a fundamental property of retrieval-oriented embeddings, our work encourages the research community to investigate and develop new privacy-preserving representation learning techniques that explicitly account for semantic inference risks.

The potential negative impacts include the misuse of the proposed analysis framework to extract sensitive information from embedded systems, such as retrieval-augmented generation and vector database. However, we emphasize that the primary goal of this work is to expose systemic vulnerabilities so that they can be discussed and addressed proactively.

Overall, this work contributes to more secure and trustworthy deployment of embeddings by motivating semantic-level privacy analysis and mitigation as a core consideration in representation learning.




\nocite{langley00}

\bibliography{references}
\bibliographystyle{icml2026}

\newpage
\appendix
\onecolumn

\section{Derivation of Normal Equation}~\label{normal_equation}
To determine the optimal alignment matrix $\mW$, we aim to minimize a cost function $J$ that quantifies the discrepancy between the attack embedding matrix $\mE_{A}$ and the transformed victim embeddings $\mE_{V\rightarrow A}=\mE_{V} \mW$: 

\begin{equation}
\begin{aligned}
J(\mW) &= \frac{1}{2}\,\mathrm{tr}\!\left((\mE_A  - \mE_V \mW)^{T} (\mE_A - \mE_V \mW)\right) \\ 
& = \frac{1}{2}\,\mathrm{tr}\!\left(\mE_A^{T} \mE_A - \mE_A^{T} \mE_V \mW -  (\mE_{V} \mW)^{T} \mE_{A}  + (\mE_{V} \mW)^{T} \mE_V \mW\right) \\ 
& = \frac{1}{2}\,\mathrm{tr}\!\left(\mE_A^{T} \mE_A - \mE_A^{T} \mE_V \mW -   \mW^{T}\mE_{V}^{T} \mE_{A}  +  \mW^{T}\mE_{V}^{T} \mE_V \mW\right). 
\end{aligned}
\end{equation}
By calculating the derivatives of $J(\mW)$, we have
\begin{equation}
    \begin{aligned}
        \nabla_{\mW} J(\mW)  & =\frac{1}{2} \nabla_{\mW} \mathrm{tr}\!\left(\mE_A^{T} \mE_A - \mE_A^{T} \mE_V \mW  - \mW^{T}\mE_{V}^{T} \mE_{A} +  \mW^{T}\mE_{V}^{T} \mE_V \mW\right) \\ 
        & = \mE_{V}^{T} \mE_V \mW - \mE_{V}^{T} \mE_{A} .
    \end{aligned}
\end{equation}
The optimized $\mW$ is achieved when the derivative is equal to 0,
\begin{equation}
    \mE^{T}_{V} \mE_{V} \mW = \mE^{T}_{V} \mE_{A}.
\end{equation}
Then, the matrix $\mW$ that minimizes $J(\mW)$ is (assuming $\mE_V^\top \mE_V$ is invertible)
\begin{equation}
\mW = (\mE_V^{T}\mE_V)^{-1}\mE^{T}_{V} \mE_{A}.
\end{equation}



\section{Models}\label{app:models}

\begin{table}[h!]
    \centering
        
    \caption{Details of the Models}
\resizebox{0.8\linewidth}{!}{
    \begin{tabular}{c|cccccc}
    \toprule 
        Model       &  Usage & Dimension & \# Parameters &  Platform &  Access  & Name\\ 
        \midrule
\textsc{Nomic}  & Multimodal Embedder & 768 & 92.9M& Huggingface &Open-Source & nomic-ai/nomic-embed-vision-v1.5 \\
\textsc{Clip}  & Multimodal Embedder & 768 & 426 &Huggingface & Open-source& openai/clip-vit-large-patch14\\
       \textsc{Gemini}  & Multimodal Embedder & 1408 &  - & Vertex AI& Proprietary & multimodalembedding@001\\
\textsc{Cohere}  & Multimodal Embedder & 1536 & - & Cohere & Proprietary & embed-v4.0 \\
\midrule
\textsc{Gpt-5.2}  & LLM &- & - &  OpenRouter & Proprietary &openai/gpt-5.2-chat \\
\textsc{Deepseek-v3.2}  & LLM &- &  685 B &  OpenRouter &Open-source & deepseek/deepseek-v3.2\\
\textsc{Qwen3-235B}  & LLM & - &  235B &OpenRouter&Open-source & qwen/qwen3-235b-a22b-2507\\
\textsc{Cohere v2}  & LLM & -&111B  &Cohere &Proprietary &  command-a-03-2025\\
\textsc{Minimax-m1}  & LLM &- & 456B &Huggingface &  Open-source & MiniMaxAI/MiniMax-M1-80k\\
\midrule 
\textsc{Gemini-flash}  & VLM & -& - & OpenRouter& Proprietary & google/gemini-3-flash-preview\\
\textsc{Gpt-5.1}  & VLM & -& - &OpenRouter&  Proprietary& openai/gpt-5.1-chat\\
\textsc{Qwen2.5-vl}  & VLM & -& 72B & OpenRouter&  Open-source&  Qwen/Qwen2.5-VL-72B-Instruct\\
\midrule 
\textsc{Kandinsky 2.2}  & Stable Diffusion Model& - & 271B & Huggingface & Open-Source &  kandinsky-community/kandinsky-2-2-decoder \\

\textsc{CLIP} in \textsc{Kandinsky 2.2}   & Dual Encoder & 1280 & 1.8B & Huggingface & Open-Source &  ViT-bigG-14-laion2B-39B-b160\\
       \bottomrule
    \end{tabular}}
    \label{tab:model_detail}
\end{table}

\section{Statistics of Processed Tags of \textsc{COCO} Training Data}\label{stats_app}
\begin{figure}
    \centering
    \includegraphics[width=0.85\linewidth]{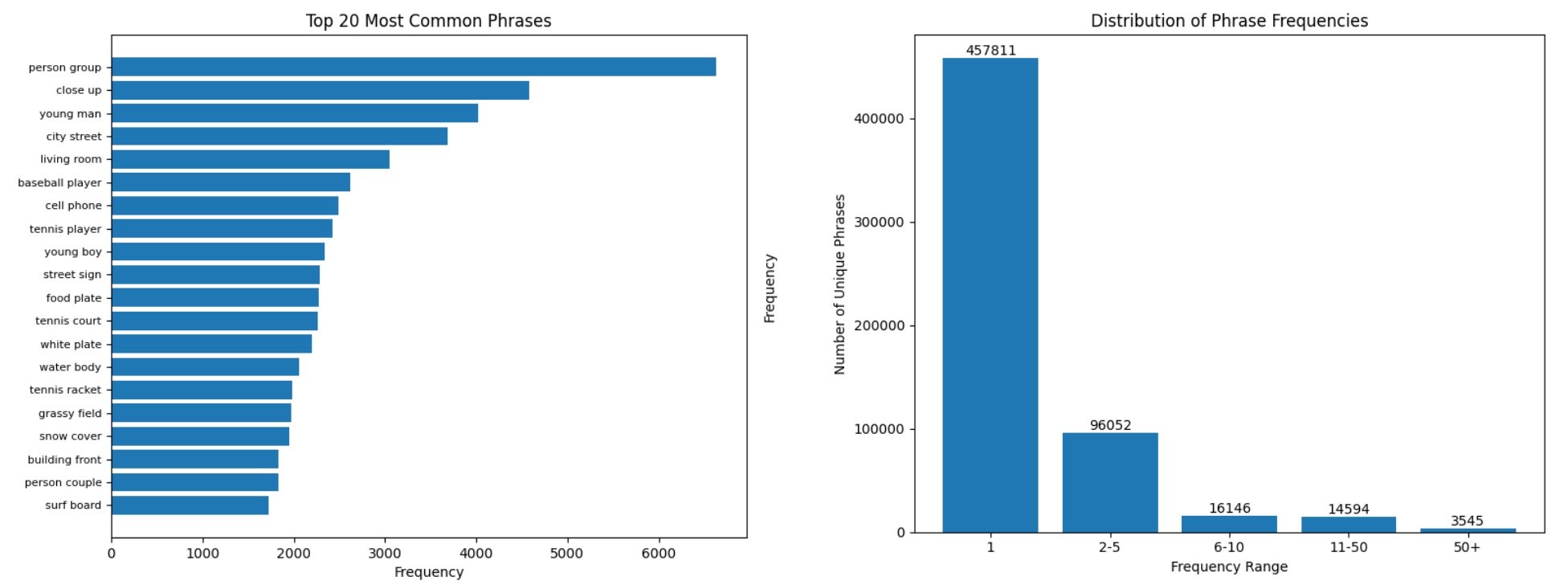}
    \caption{(Left) Top 20 Most Common Tag Phrases and their Frequencies; (Right) Distribution of Unique Tags. }
    \label{fig:distribution_tags}
\end{figure}

We have 122,285 training data samples, totaling 584,050 captions. 
After preprocessing the tags from the captions, there are in total 1,750,750 relation tags, on average 14.3 for each image.
We have in total $91,018$ triples (relations) and $41,654$ unique ones.
Fig.~\ref{fig:distribution_tags} shows distributions of triples and unique tags.

\newpage
\section{Evaluate LLM-Generated Texts}\label{eval_llm_text}
An image can convey multiple meanings and can therefore be associated with multiple textual descriptions. 
Evaluating $n$-to-$n$ correspondences between human-written and human-generated texts requires metrics that capture complementary aspects of privacy risk, quality, and overlap.

Let $I=\{i_1, \cdots, i_n\}$ be a set of images. 
For each image $i$, let $R_i=\{r_{i,1}, \cdots, r_{i,K}\}$ be the set of human-written reference texts. 
Let $H_i=\{h_{i,1},\cdots, h_{i,M}\}$ be the set of LLM-generated hypotheses.
Let $s(h, R)$ denote a reference-based scoring function, such as BLEU, METEOR, Rouge-L, that compares a hypothesis $h$ against a set of references $R$.
Let $s(h, r)$ denote the same metric applied to a single reference. 


Each hypothesis is scored against its best-matching references, i.e., $\text{Score}_{best}= \frac{1}{N}\sum^{N}_{i=1}(\frac{1}{M} \sum^{M}_{j=1} \max_{k\in \{1,\cdots,{K}\}} s(h_{i,j}, r_{i,k}))$.
This captures the worst-case privacy leakage, where a single accurate reconstruction suffices, and is most suitable for evaluating the vulnerability of embeddings in our setting.

\begin{figure}
    \centering
        \includegraphics[width=0.5\linewidth]{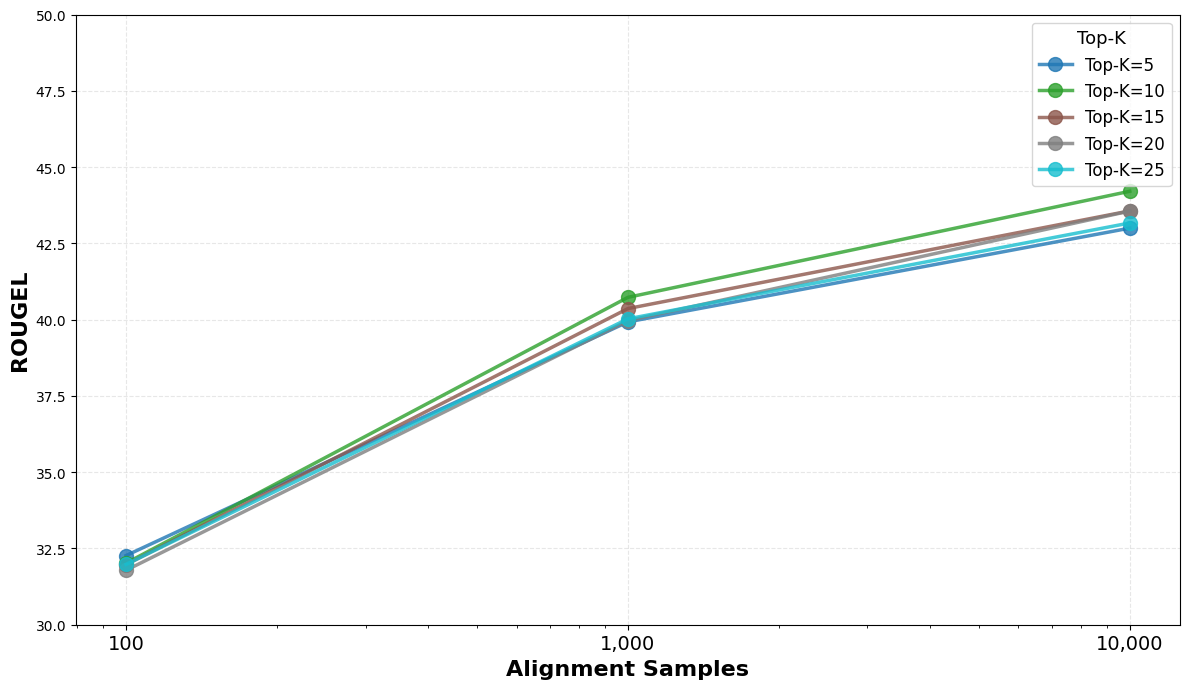}
            \caption{Ablation Study of $K$, Evaluated against $\{C_{gt}\} $ on attacking \textsc{Gemini} Embeddings across numbers of Alignment Samples.}
                \label{fig:ablations_gemin_topk}
\end{figure}

\begin{figure}
    \centering
    \includegraphics[width=\linewidth]{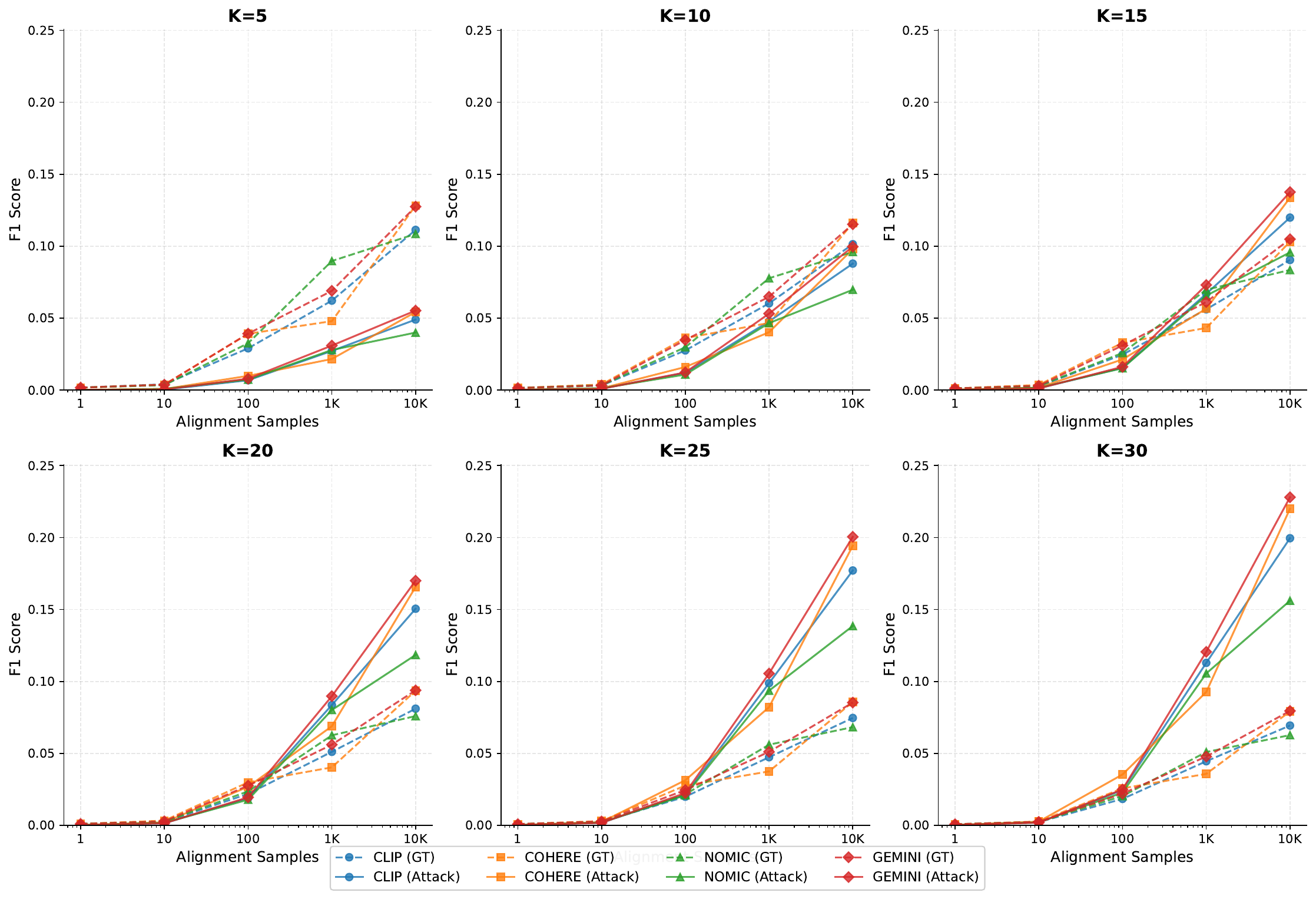}
     \caption{Retrieval of Exact Tags in F1 Scores across K, evaluated against  $\{t_{gt}\}$ and   $\{t_{A}\}$}.

    \label{fig:exact_tag_retrieval}
\end{figure}

\section{Results on Text Reconstruction}

\begin{figure}[h!]
    \centering
   
    \includegraphics[width=0.49\linewidth]{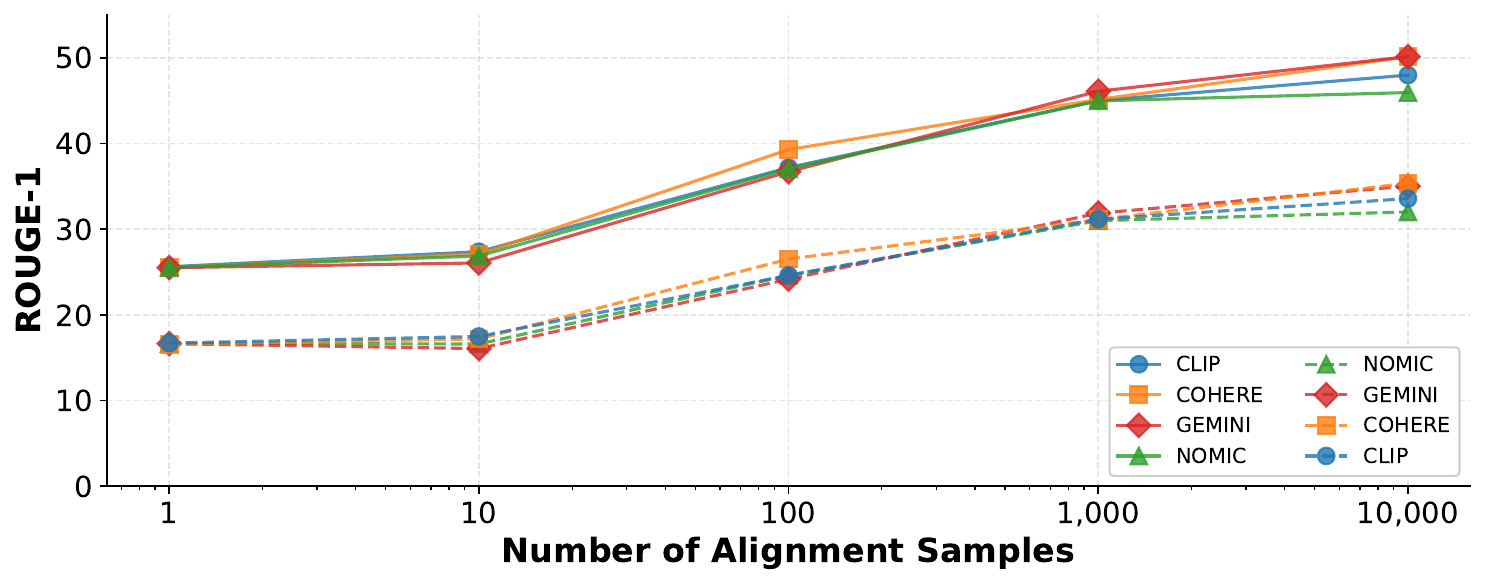}
\includegraphics[width=0.49\linewidth]{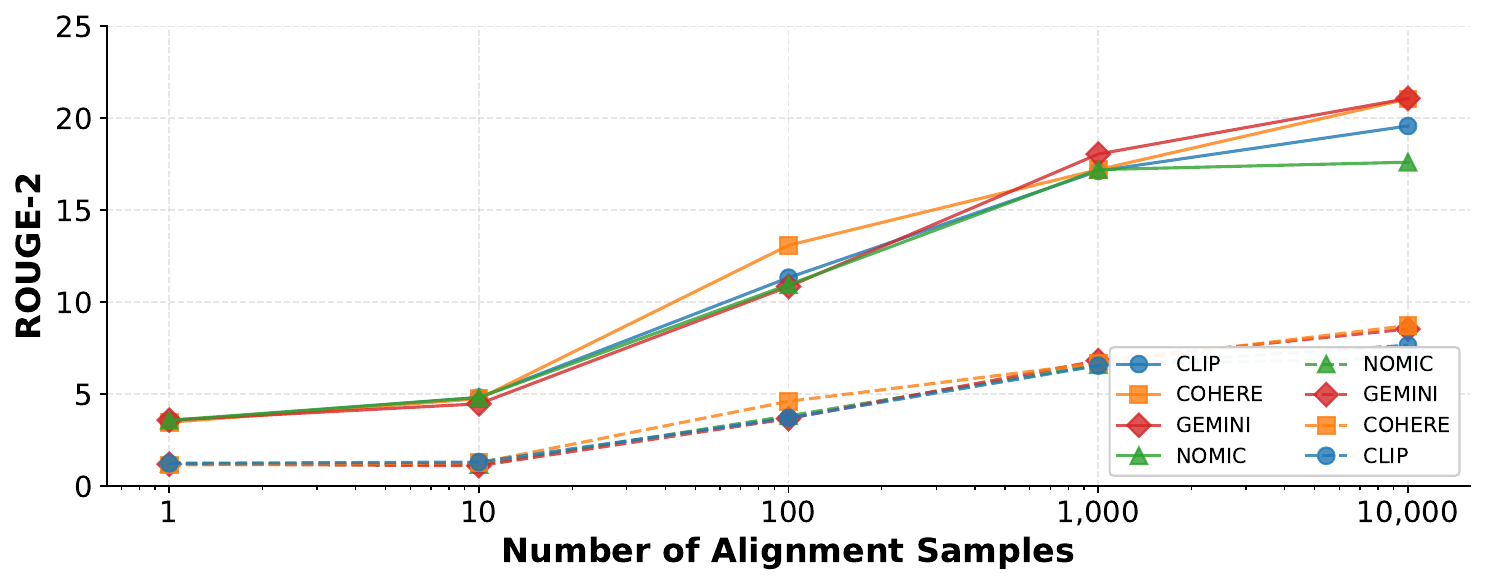}
    \includegraphics[width=0.49\linewidth]{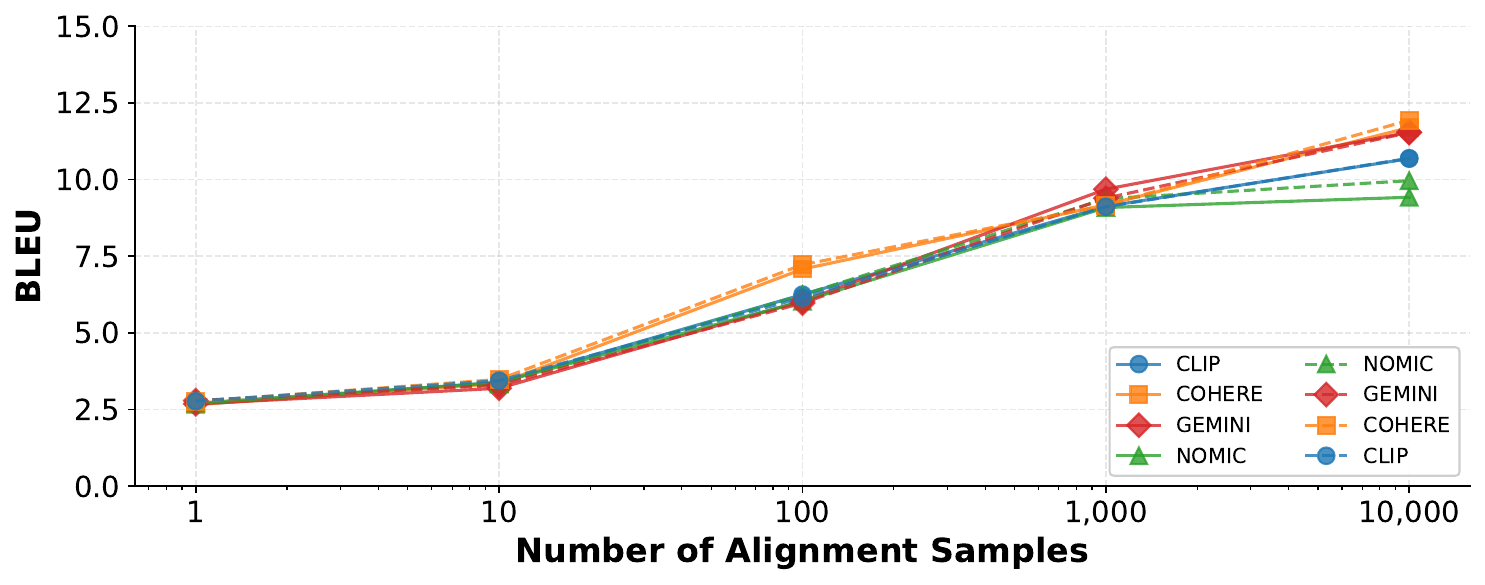}
\includegraphics[width=0.49\linewidth]{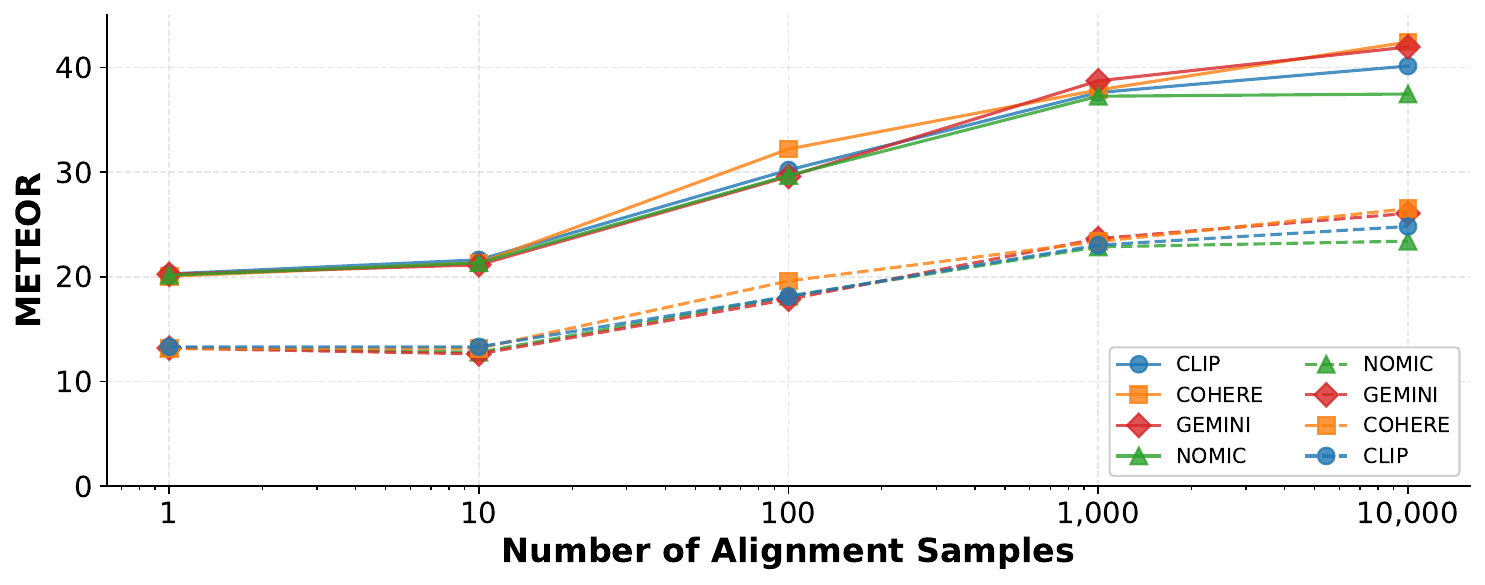}
 \caption{Semantic Leakage of Image Embeddings by Alignment Samples. $\{C_{V\rightarrow A}\}$ evaluated against $\{C_{gt}\}$ (solid lines) and $\{C_{h}\}$ (dotted lines) against at $K=10$ in ROUGE-1, ROUGE-2, BLEU and METEOR.}
    \label{fig:results_full_text_reconstruction}
\end{figure}

\newpage
\section{Adaptive Attacks}

\subsection{Relations for Adaptive Attacks}\label{relations_app}
For relation-level evaluation, we restrict the predicate vocabulary to a fixed set of common spatial and part-whole relations:
\{\texttt{on}, \texttt{over}, \texttt{under}, \texttt{inside}, \texttt{covering}, \texttt{hanging\_over}, \texttt{enclosing}, \texttt{next\_to}, \texttt{part\_of}\}.

Prior work has shown that a small number of spatial and compositional relations account for the majority of predicate instances in large-scale scene graph datasets~\citep{krishna2017visual,zellers2018neural}.
This set is widely adopted in scene graph generation to aid evaluation~\citep{xu2017scene}.

\begin{figure}[h!]
    \centering
    \includegraphics[width=0.49\linewidth]{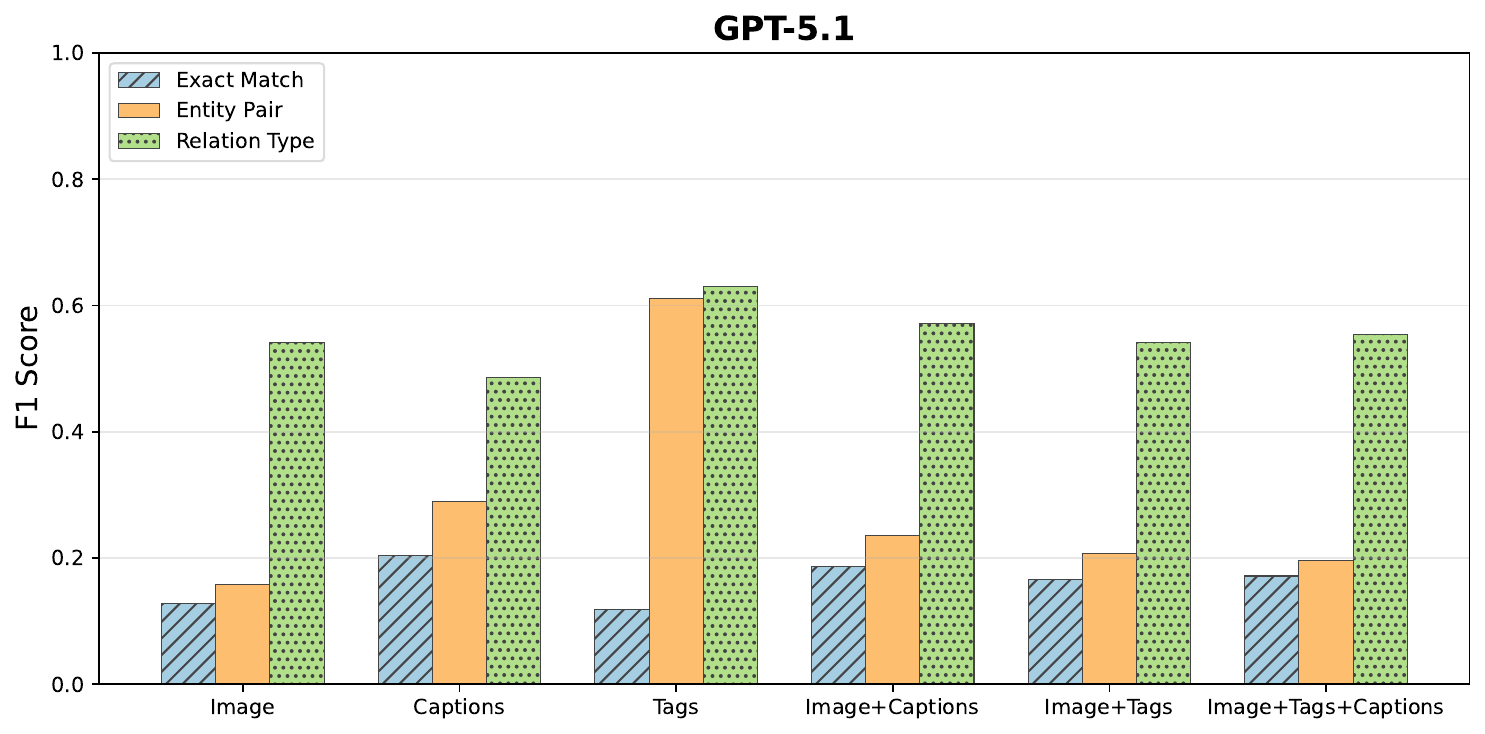}
    \includegraphics[width=0.49\linewidth]{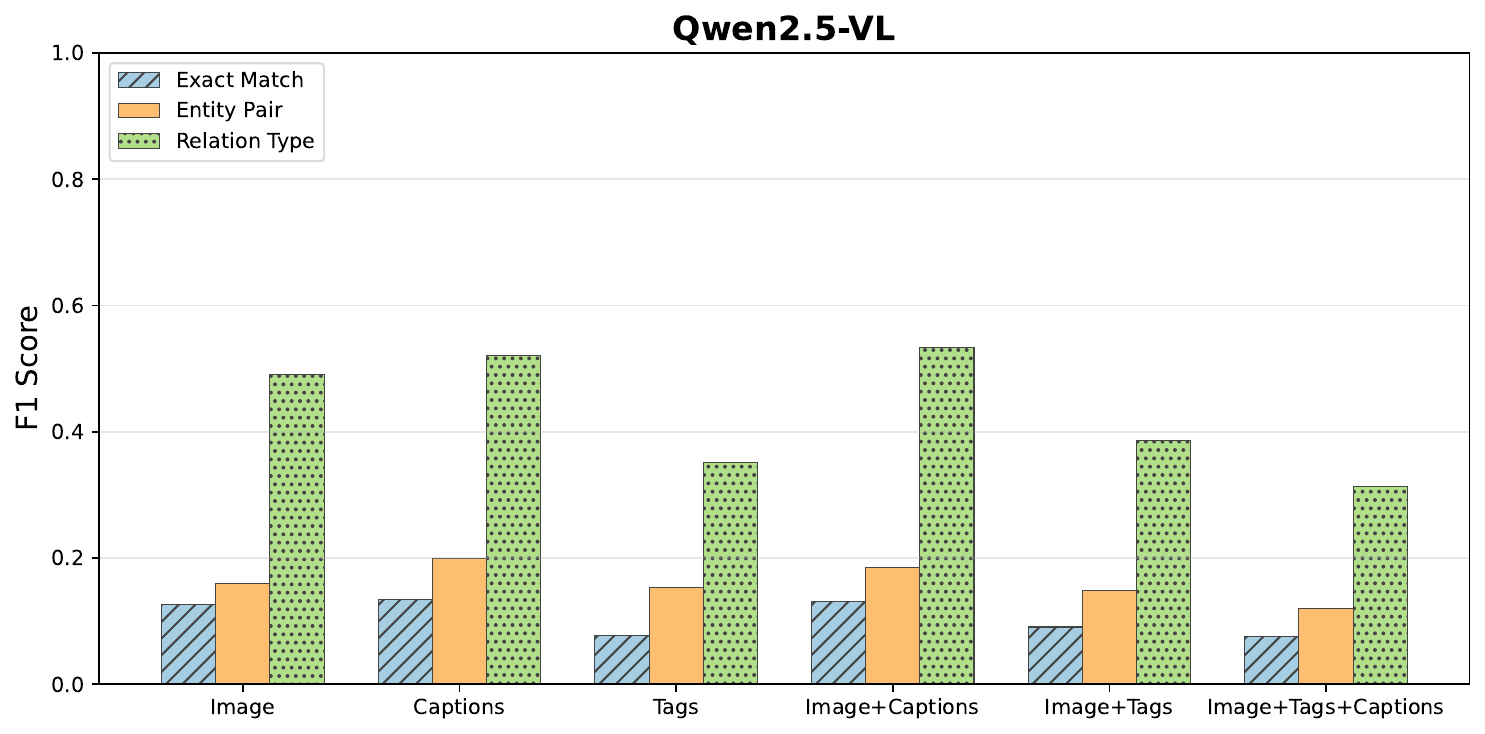}
        \caption{Identifying Relations across Settings in F1.}

    \label{fig:qwenvl_gpt_relations}
\end{figure}

\begin{figure}[h!]
    \centering
    \includegraphics[width=0.49\linewidth]{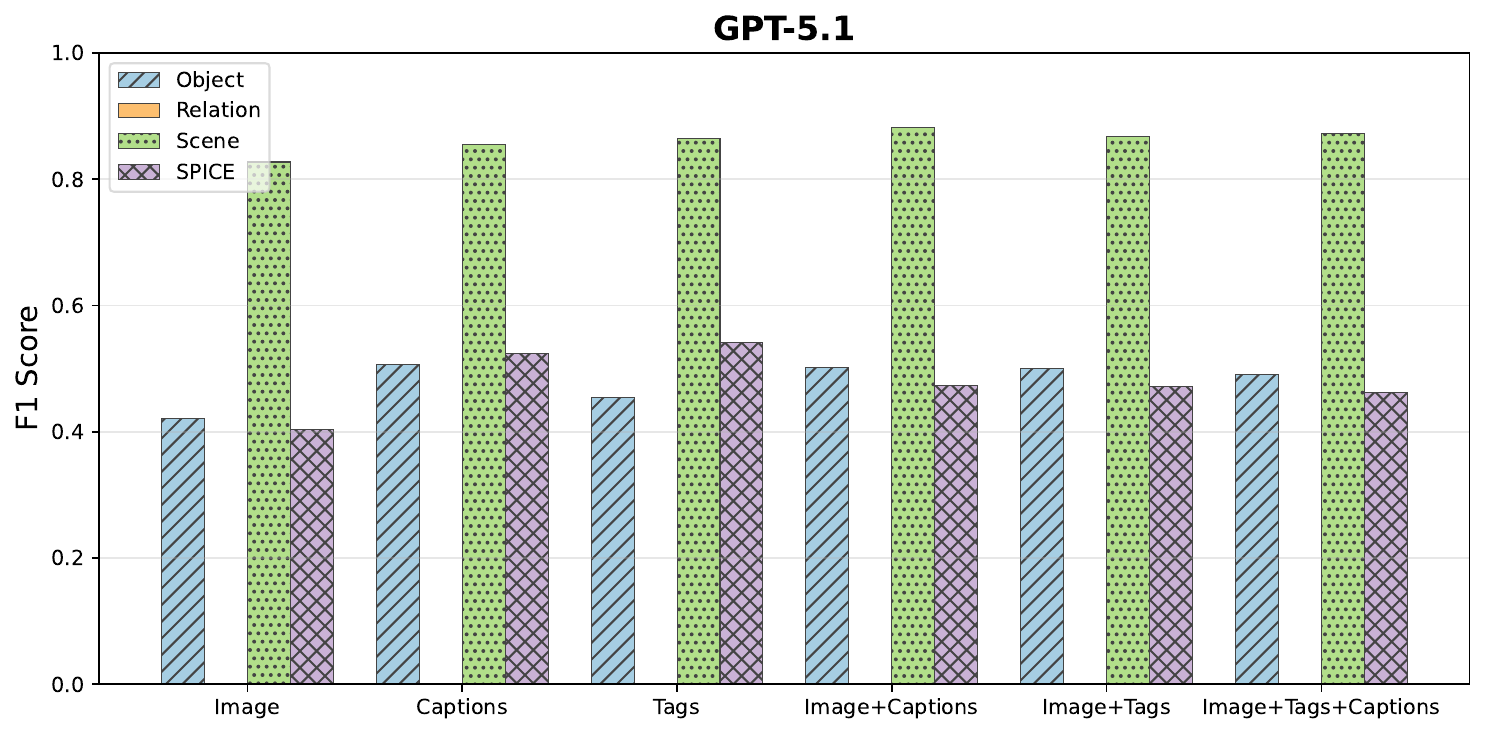}
    \includegraphics[width=0.49\linewidth]{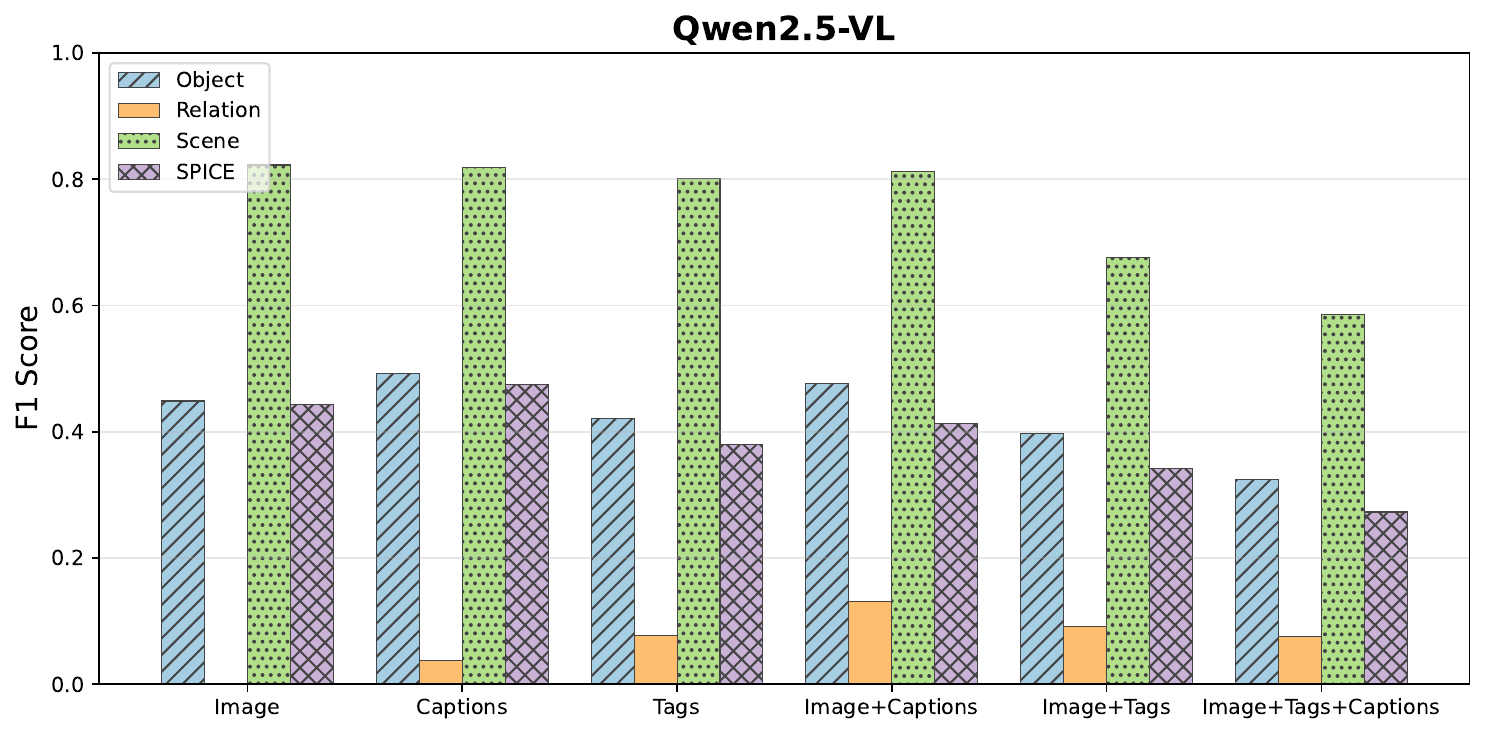}
         \caption{Identifying Scene Graphs across Settings in F1.}

    \label{fig:qwenvl_gpt_scene_graphs}
\end{figure}

\subsection{Full Results in Ablation Studies}\label{sec:ablation_app}
\begin{figure}
    \centering
        \includegraphics[width=0.5\linewidth]{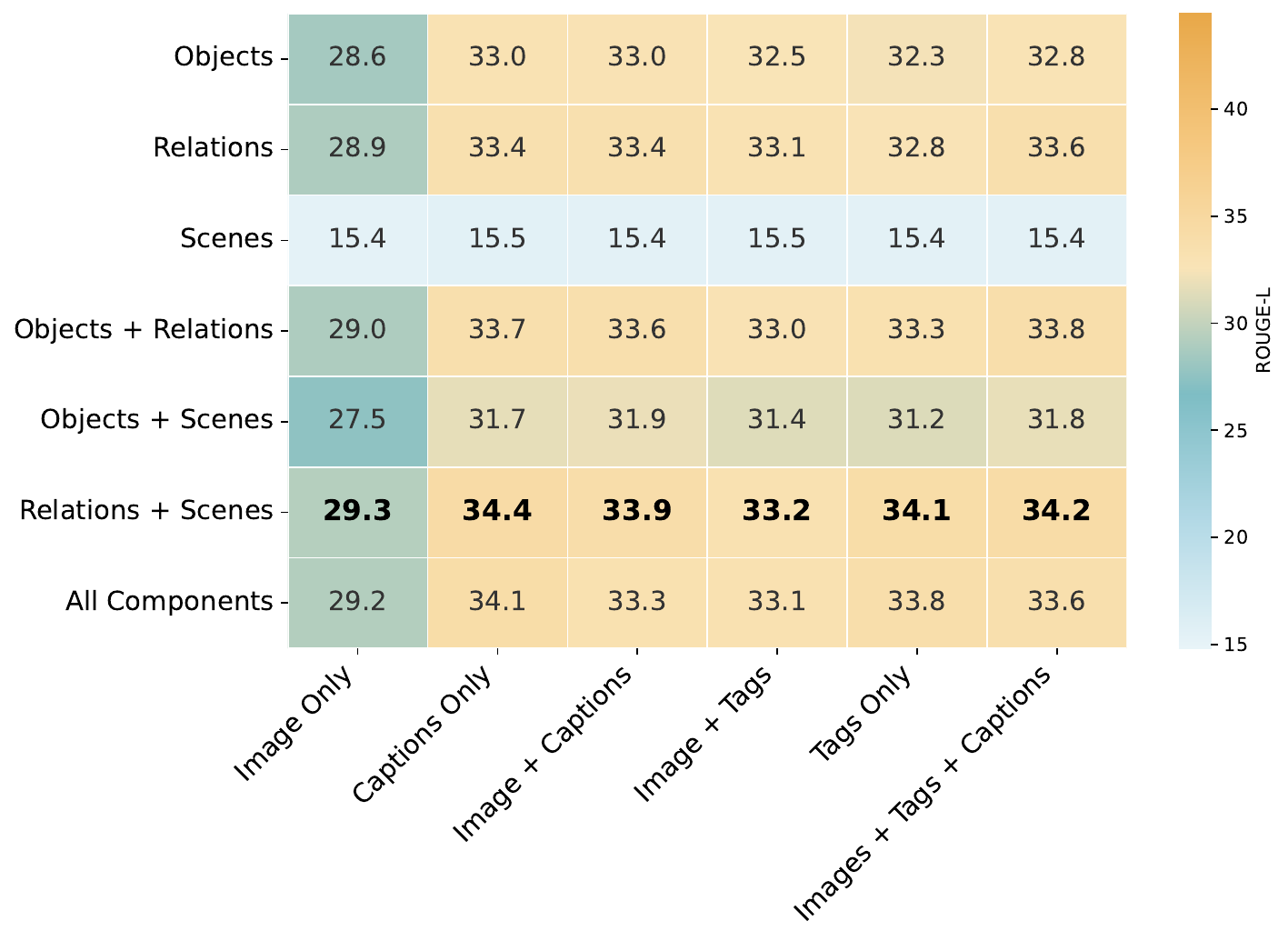}
             \caption{Evaluate Generated Captions \(\{C_{V \rightarrow A}\} \) against \(\{C_{h}\}\) across settings, the best Rouge-L score is \textbf{bolded} for each column.}
    \label{fig:ablation_study_rougel_gt}
\end{figure}

\begin{figure}
    \centering
    \includegraphics[width=0.49\linewidth]{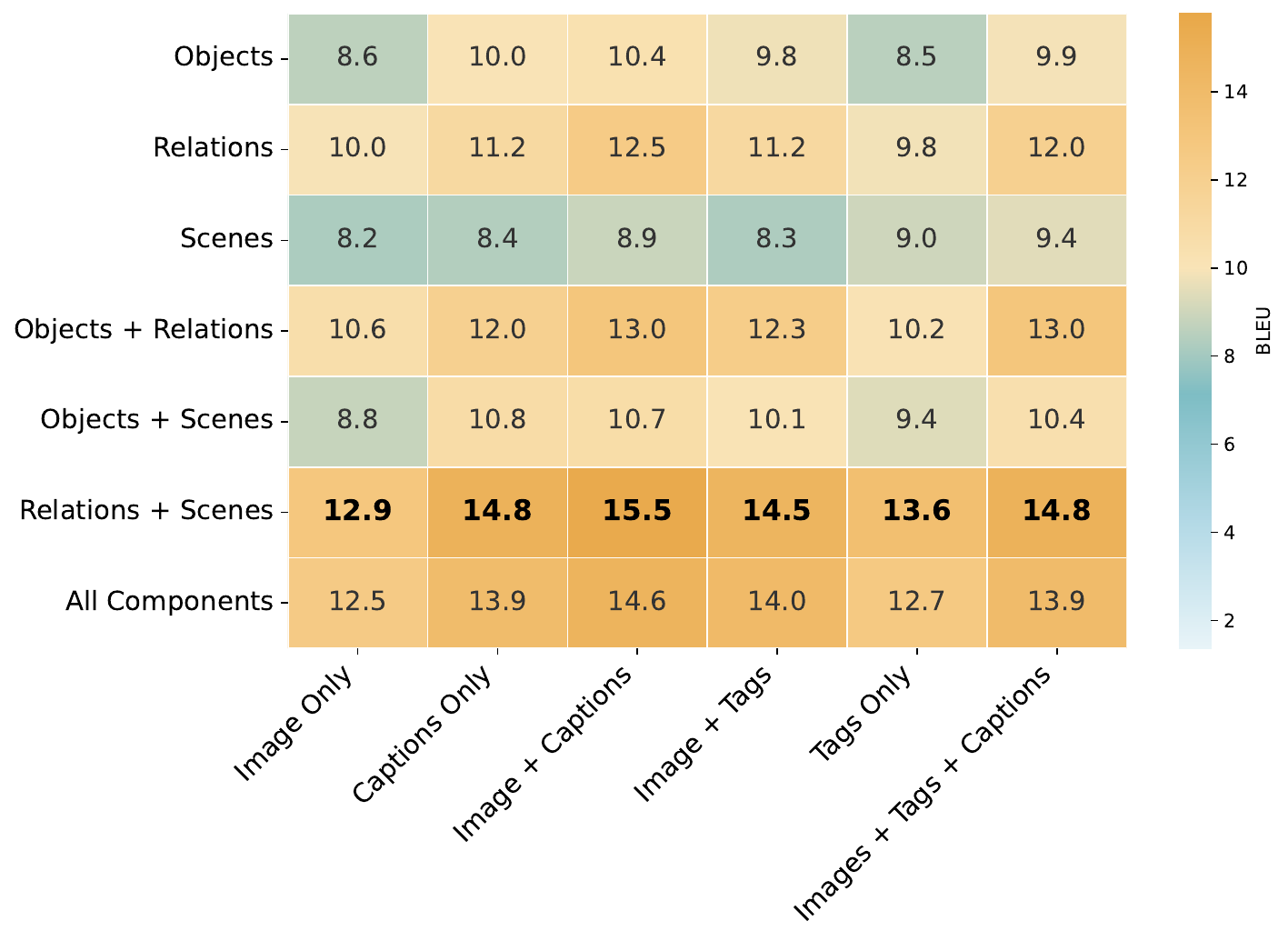}
    \includegraphics[width=0.49\linewidth]{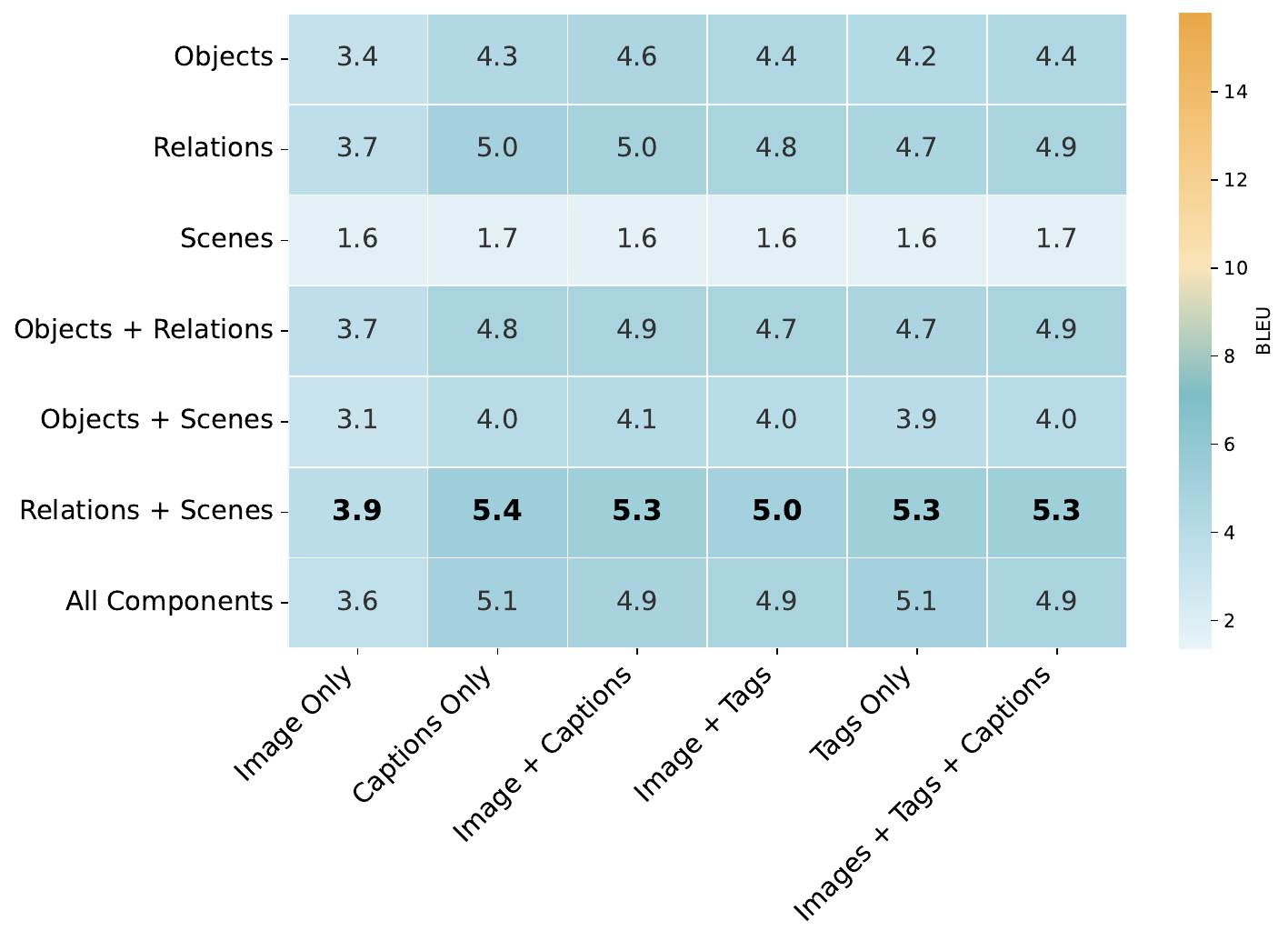}
    \caption{Evaluate Generated Captions \(\{C_{V \rightarrow A}\} \) against (Left) \(\{C_{gt}\}\) and (Right)\(\{C_h\}\) across settings in BLEU-4, the best score is \textbf{bolded} for each column. }

    \label{fig:ablation_bleu}
\end{figure}

\begin{figure}
    \centering
    \includegraphics[width=0.49\linewidth]{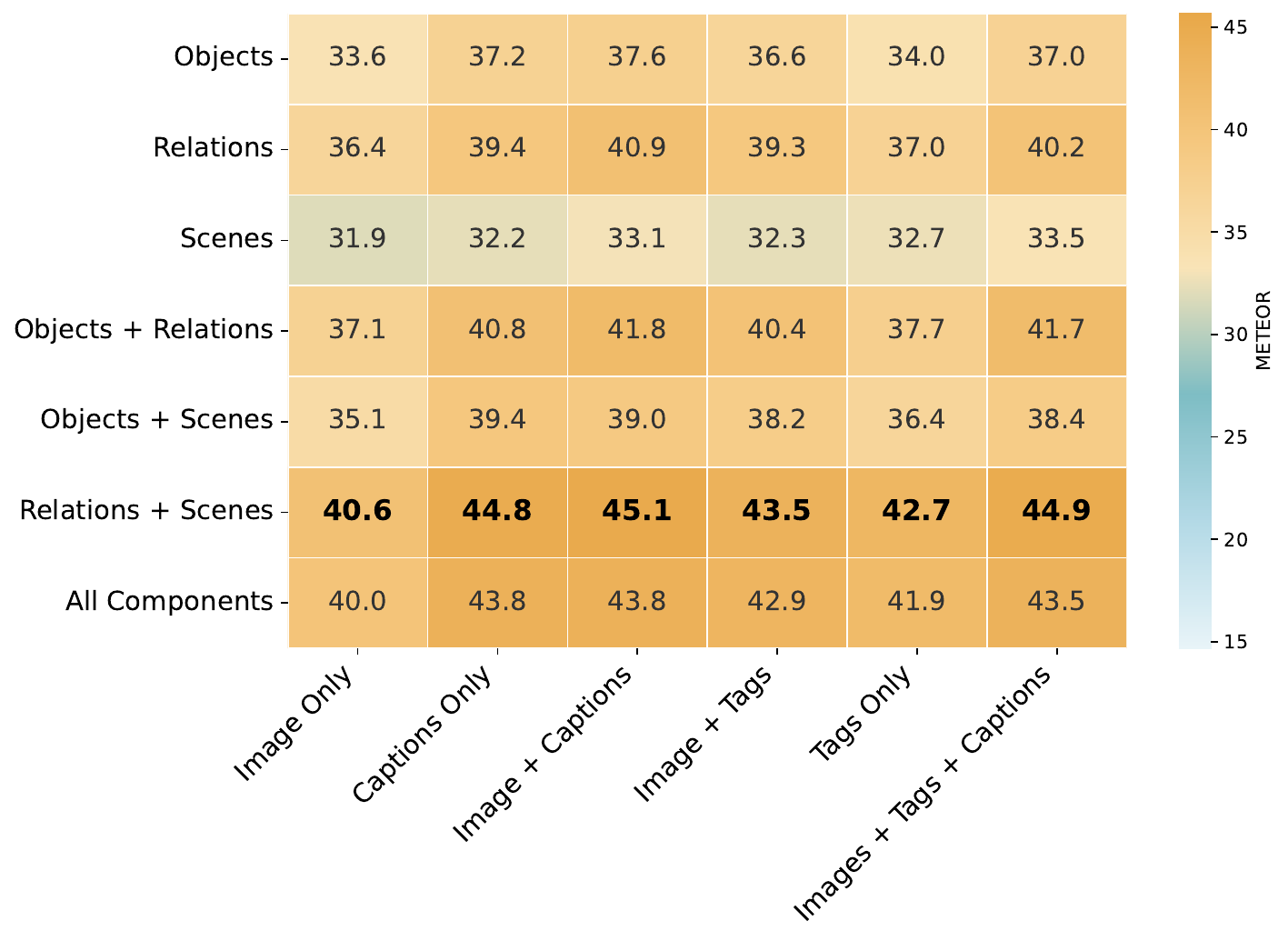}
    \includegraphics[width=0.49\linewidth]{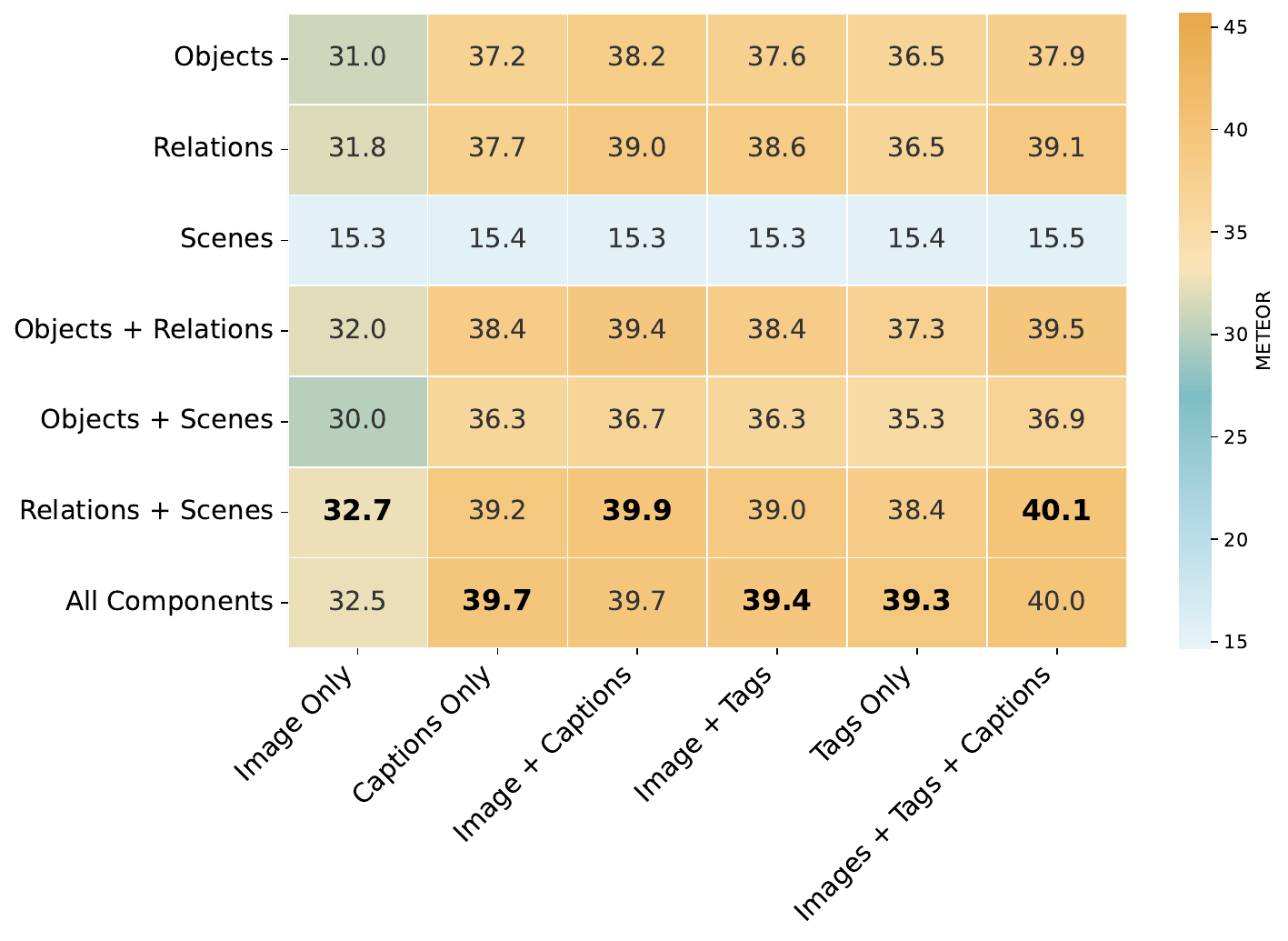}
        \caption{Evaluate Generated Captions \(\{C_{V \rightarrow A}\} \) against (Left) \(\{C_{gt}\}\) and (Right)\(\{C_h\}\) across settings in METEOR, the best score is \textbf{bolded} for each column.}

    \label{fig:ablations_meteor}
\end{figure}

\newpage

\section{Cross-Domain Evaluation}\label{cross-domian_appendix}

\paragraph{Tags Overlapping}

We extract relational tags first from the human-annotated captions from \textsc{nocaps}, and then evaluate the retrieved tags from attacked image embeddings encoding the images from \textsc{nocaps}.
Table~\ref{tab:cross_domain_tag} further validates \ourattack's generalizability across-domain data.

\begin{table}[!h]
    \centering
    \caption{Cross-Domain Evaluation on Retrieved Tags  \(\{t_{V\rightarrow A}\}\) against Ground Truth Tags \(\{t_{gt}\}\), when $K=10$. The best Rouge-L scores are \textbf{bolded}.}
      \resizebox{0.6\linewidth}{!}{
    \begin{tabular}{cccccc}
    
    \toprule
          \textbf{Victim Embedder} & \textbf{BLEU-4} & \textbf{ROUGE-1} & \textbf{ROUGE-2} & \textbf{ROUGE-L} & \textbf{METEOR} \\
        \midrule
         \textbf{Near-Domain} & & & & &  \\
        
clip & 13.14    & 46.87          & 11.25        & 45.49     & 33.26  \\
cohere   & 13.58    & 48.71           & 12.12       & \textbf{47.13}       & 34.56    \\
gemini   & 13.46     & 48.09         & 11.99    & 46.67    & 34.25    \\
nomic    & 11.85       & 42.68        & 8.46        & 41.54     & 29.24         \\
 \midrule
      \textbf{Out-Domain}& & & & & \\
clip     & 11.49                    & 41.34                      & 7.33                       & 40.36                      & 27.45                      \\
cohere   & 12.35                    & 44.45               & 8.86                       & \textbf{43.33}                     & 30.08                      \\
gemini   & 12.05                    & 43.49                      & 8.37                       & 42.36                      & 29.40                      \\
nomic    & 9.63                     & 34.72                      & 4.92                       & 34.07                      & 22.86            \\        

       \bottomrule
    \end{tabular}}
    \label{tab:cross_domain_tag}
\end{table}

\newpage

\section{Input \& Output via \ourattack}\label{app:sample}
We show a sample from \textsc{COCO} test dataset, illustrate the inputs and outputs via \ourattack alignment and inferences.

\begin{figure} [h!]
    \centering    
    \begin{subfigure}{0.35\linewidth}
    \centering
    \caption{Original Image}
        \includegraphics[width=\linewidth]{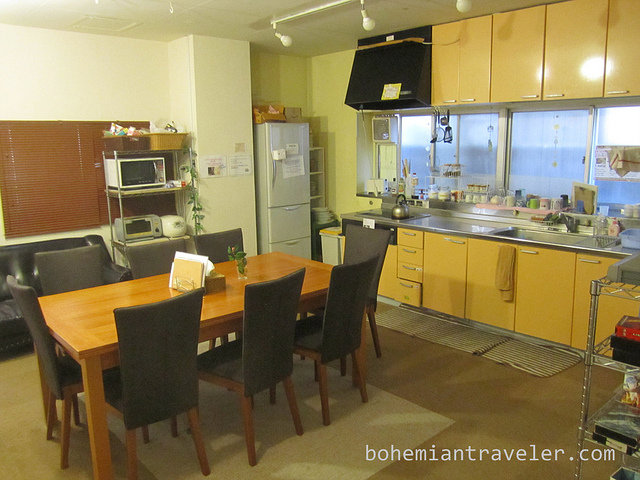}
        \label{fig:original_image}
    \end{subfigure}
    \quad
    \begin{subfigure}{0.3\linewidth}
    \centering
    \caption{ Low-fidelity Reconstruction }
    
        \includegraphics[width=\linewidth]{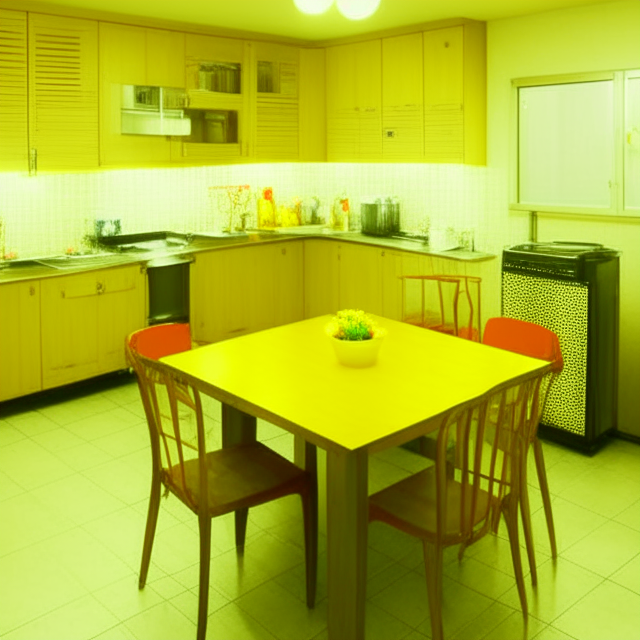}
        \label{fig:lofi_reconstruction}
    \end{subfigure}
        \caption{(a) An original image sample from \textsc{COCO} test dataset, and (b) diffusion-model generated image directly from its aligned \textsc{Gemini} image embedding.}

    \label{fig:image_sample}
\end{figure}

\begin{tcolorbox}[
    title={$\{C_h\}$: Human-written Captions for Image in Fig.~\ref{fig:original_image}},
    colback=gray!5,
    colframe=black!50,
    arc=4pt,
    fonttitle=\bfseries,
    boxrule=0.5pt
]
The kitchen area is clean and ready for us to use.

A kitchen with a wooden table surrounded by black chairs.

A small kitchen and dining room area that has multiple metal shelves with items on it.

The kitchen and eating area of an apartment.

A kitchen and dining area combined into one open space with track lighting on the ceiling.
\end{tcolorbox}

\begin{tcolorbox}[
    title={$\{t_{gt}\}$ Processed tags from $\{C_h\}$},
    colback=gray!5,
    colframe=black!50,
    arc=4pt,
    fonttitle=\bfseries,
    boxrule=0.5pt
]
multiple shelf, kitchen area, apartment area, area combine into space, metal shelf, wooden table, track lighting, open space, table kitchen, room area, multiple metal shelf, that have shelf, area combine with lighting, dining room, it item, area combine on ceiling, black chair, small area, kitchen apartment area, small kitchen room area, that have with item
\end{tcolorbox}

\begin{tcolorbox}[
  title={$\{C_{gt}\}$ \textsc{Deepseek v3.2}-Generated Captions from $\{t_{gt}\}$},
  colback=gray!5,
    colframe=black!50,
    arc=4pt,
    fonttitle=\bfseries,
    boxrule=0.5pt
]
The kitchen and dining room combine into one open space with multiple metal shelves. 

A small apartment area has track lighting on the ceiling. 

A wooden table and black chairs are in the kitchen area. 

Multiple shelves hold various items in the small kitchen room. 

The open space combines a kitchen and dining area with a wooden table.
\end{tcolorbox}

\begin{tcolorbox}[
  title={$\{t_{A}\}$ from attack embeddings $\ve_A$ encoded by \textsc{Gemini} embedder ($K=10$)},
  colback=gray!5,
    colframe=black!50,
    arc=4pt,
    fonttitle=\bfseries,
    boxrule=0.5pt
]
kitchen table, table kitchen, yellow kitchen, wooden kitchen table, wood kitchen table, kitchen chair, table space, empty table kitchen, kitchen feature table, table have kitchen

\end{tcolorbox}

\begin{tcolorbox}[
  title={Captions generated from Relations and Scenes in Fig~\ref{fig:objects_relations_scenes_result_sample_gt} using \textsc{Gemini-Flash}},
  colback=gray!5,
    colframe=black!50,
    arc=4pt,
    fonttitle=\bfseries,
    boxrule=0.5pt
]
In this organized kitchen, a microwave and toaster oven sit on a shelf located next to the refrigerator. 

A table with chairs and a sofa is illuminated by lighting hanging over it, situated near a window with blinds. 

The functional kitchen area features cabinets mounted over a sink, with a storage shelf placed next to the window blinds. 

A cozy living space transitions into a kitchen where a lighting fixture hangs over a table positioned next to a sofa. 

The kitchen layout includes a sink next to a cabinet, while various appliances rest on a shelf beside the refrigerator.

\end{tcolorbox}

\begin{figure}[t]
\centering
\footnotesize
\begin{tcolorbox}[
  colback=gray!3,
  colframe=black!40,
  arc=3pt,
  boxrule=0.4pt,
  title={Adaptive Attack Results from Image in Fig.~\ref{fig:original_image} and Captions \(\{C_{h}\}\) for Evaluation Baseline},
  fonttitle=\bfseries
]
\noindent
\begin{minipage}[t]{0.33\linewidth}
\vspace{0pt}
\textbf{Objects}
\begin{itemize}[leftmargin=*, nosep]
  \item table (\emph{multiple})
  \item chair (\emph{multiple})
  \item shelf (\emph{multiple})
  \item cabinet (\emph{image})
  \item refrigerator (\emph{image})
  \item microwave (\emph{image})
  \item sink (\emph{image})
  \item toaster oven (\emph{image})
  \item lighting (\emph{multiple})
  \item sofa (\emph{image})
  \item window (\emph{image})
  \item blind (\emph{image})
\end{itemize}
\end{minipage}\hfill
\begin{minipage}[t]{0.47\linewidth}
\vspace{0pt}
\textbf{Relations}
\begin{itemize}[leftmargin=*, nosep]
  \item chair \(\xrightarrow{\texttt{next\_to}}\) table (\emph{multiple})
  \item microwave \(\xrightarrow{\texttt{on}}\) shelf (\emph{image})
  \item toaster oven \(\xrightarrow{\texttt{on}}\) shelf (\emph{image})
  \item shelf \(\xrightarrow{\texttt{next\_to}}\) refrigerator (\emph{image})
  \item sink \(\xrightarrow{\texttt{next\_to}}\) cabinet (\emph{image})
  \item cabinet \(\xrightarrow{\texttt{over}}\) sink (\emph{image})
  \item sofa \(\xrightarrow{\texttt{next\_to}}\) table (\emph{image})
  \item blind \(\xrightarrow{\texttt{covering}}\) window (\emph{image})
  \item lighting \(\xrightarrow{\texttt{hanging\_over}}\) table (\emph{multiple})
  \item shelf \(\xrightarrow{\texttt{next\_to}}\) blind (\emph{image})
\end{itemize}
\end{minipage}\hfill
\begin{minipage}[t]{0.18\linewidth}
\vspace{0pt}
\textbf{Scenes}
\begin{itemize}[leftmargin=*, nosep]
  \item kitchen (0.95, \emph{multiple})
  \item living room (0.40, \emph{image})
\end{itemize}
\end{minipage}
\end{tcolorbox}

\caption{Representative structured output for one sample. Parentheses indicate confidence (for scenes) and the supporting modality (\emph{image} vs.\ \emph{multiple}).}

\label{fig:objects_relations_scenes_result_sample_gt}
\end{figure}

\begin{figure}[t]
\centering

\footnotesize
\setlength{\tabcolsep}{4pt}

\begin{tcolorbox}[colback=gray!3,colframe=black!40,arc=3pt,boxrule=0.4pt,
  title={Adaptive Attack Results from Image in Fig.~\ref{fig:lofi_reconstruction} and Captions \(\{C_{V\rightarrow A}\}\)}, fonttitle=\bfseries]
\begin{minipage}[t]{0.32\linewidth}
\vspace{0pt}
\textbf{Objects}
\begin{itemize}[leftmargin=*, nosep]
  \item table (\emph{multiple})
  \item chair (\emph{image})
  \item cabinet (\emph{image})
  \item countertop (\emph{multiple})
  \item window (\emph{image})
  \item potted plant (\emph{image})
  \item stove (\emph{image})
  \item kitchen sink (\emph{image})
  \item light fixture (\emph{image})
\end{itemize}
\end{minipage}\hfill
\begin{minipage}[t]{0.44\linewidth}
\vspace{0pt}
\textbf{Relations}
\begin{itemize}[leftmargin=*, nosep]
  \item chair \(\xrightarrow{\texttt{next\_to}}\) table (\emph{image})
  \item potted plant \(\xrightarrow{\texttt{on}}\) table (\emph{image})
  \item cabinet \(\xrightarrow{\texttt{over}}\) countertop (\emph{image})
  \item stove \(\xrightarrow{\texttt{part\_of}}\) countertop (\emph{image})
  \item kitchen sink \(\xrightarrow{\texttt{part\_of}}\) countertop (\emph{image})
  \item light fixture \(\xrightarrow{\texttt{hanging\_over}}\) table (\emph{image})
  \item cabinet \(\xrightarrow{\texttt{next\_to}}\) window (\emph{image})
\end{itemize}
\end{minipage}\hfill
\begin{minipage}[t]{0.20\linewidth}
\vspace{0pt}
\textbf{Scene}
\begin{itemize}[leftmargin=*, nosep]
  \item kitchen (1.0, \emph{multiple})
\end{itemize}
\end{minipage}
\end{tcolorbox}
\caption{A representative sample of the intermediate structured representations extracted in our adaptive attack. Parentheses indicate the supporting modality (\emph{image} vs.\ \emph{multiple}).}

\label{fig:objects_relations_scenes_result_sample_test_victim_gemini}
\end{figure}

\begin{tcolorbox}[
  title={$\{t_{V \rightarrow A}\}$ from attack embeddings $\ve_V$ ($K=10$)},
  colback=gray!5,
    colframe=black!50,
    arc=4pt,
    fonttitle=\bfseries,
    boxrule=0.5pt
]
kitchen table, table kitchen, kitchen space, open kitchen, kitchen area, kitchen middle, counter space, kitchen apartment, ktichen table, apartment kitchen

\end{tcolorbox}

\begin{tcolorbox}[
  title={$\{C_{V \rightarrow A}\}$ \textsc{Deepseek v3.2}-Generated Captions from $\{t_{V \rightarrow A}\}$},
  colback=gray!5,
    colframe=black!50,
    arc=4pt,
    fonttitle=\bfseries,
    boxrule=0.5pt
]
The kitchen table is in the middle of the apartment.

This open kitchen has a lot of counter space.

The kitchen area features a large table.

An apartment kitchen with an open floor plan.

A table sits in the middle of the kitchen space.
\end{tcolorbox}

\begin{tcolorbox}[
  title={Captions generated from Relations and Scenes in Fig~\ref{fig:objects_relations_scenes_result_sample_test_victim_gemini} using \textsc{Gemini-Flash}},
  colback=gray!5,
    colframe=black!50,
    arc=4pt,
    fonttitle=\bfseries,
    boxrule=0.5pt
]
A kitchen features a stove and sink integrated into the countertop with cabinets mounted overhead and next to the window.

 A potted plant sits on the table next to a chair, directly beneath a hanging light fixture in the kitchen. 
 
The kitchen layout includes a countertop with a built-in stove and sink, complemented by several cabinets and a nearby dining table. 

In this kitchen, a light fixture hangs over a table where a potted plant rests beside a chair. 

Upper cabinets are positioned over the countertop and beside a window, overlooking a kitchen area with a small dining set.

\end{tcolorbox}


\end{document}